\pdfoutput=1

\documentclass[11pt]{article}

\usepackage{ACL2023}

\usepackage{times}
\usepackage{latexsym}
\usepackage{graphicx}
\usepackage{amsmath}
\usepackage{amssymb}
\usepackage{booktabs}
\usepackage{multirow}
\usepackage{adjustbox}
\usepackage{arydshln}
\usepackage{amsmath, amssymb, amsthm}
\usepackage{float}
\usepackage[dvipsnames]{xcolor}
\usepackage{colortbl,xcolor}  
\usepackage{xcolor} 

\definecolor{top1}{HTML}{C6EFCE} 
\definecolor{top2}{HTML}{D9E1F2} 
\definecolor{top3}{HTML}{FFF2CC} 
\newcommand{\colorboxlegend}[2]{\colorbox{#1}{\strut\hspace{0.4em}#2\hspace{0.4em}}}

\usepackage[T1]{fontenc}

\usepackage[utf8]{inputenc}

\usepackage{microtype}

\usepackage{inconsolata}

\title{Teaching Sarcasm: Few-Shot Multimodal Sarcasm Detection via Distillation to a Parameter-Efficient Student}

\author{Soumyadeep Jana \and Sanasam Ranbir Singh \\Department of Computer Science and Engineering\\
        Indian Institute of Technology Guwahati\\
         \texttt{\{sjana ,ranbir\}@iitg.ac.in}}

\begin{document}
\maketitle
\begin{abstract}
Multimodal sarcasm detection is challenging, especially in low-resource settings where subtle image-text contradictions are hard to learn due to scarce annotated data, which hinders the model's performance. Parameter-efficient fine-tuning (PEFT) methods like adapters, LoRA, and prompt tuning reduce overfitting but struggle to reach optimal performance due to limited supervision from few-shot data. We propose \textbf{PEKD}, a unified framework that enhances PEFT methods via distillation from an expert model trained on large-scale sarcasm data, which acts as the teacher. To mitigate unreliable signals from the teacher, we introduce an entropy-aware gating mechanism that dynamically adjusts the distillation strength based on teacher confidence. Experiments on two public datasets demonstrate that our PEKD framework enables PEFT methods to outperform both prior parameter-efficient approaches and large multimodal models, achieving strong results in the few-shot scenario. The framework is modular and adaptable to a wide range of multimodal models and tasks. The code is available at \url{https://github.com/mr-perplexed/kd_sarcasm}.
\end{abstract}

\section{Introduction}

The use of multimodal (image+text) sarcasm has grown rapidly on social media, as it allows the users to express their harsh opinions in a veiled manner. Creation of large sarcasm datasets for tackling this problem is both costly and challenging \cite{davidov-etal-2010-semi, GonzlezIbez2011IdentifyingSI}, as sarcasm is often context-dependent, culturally nuanced, and difficult to annotate consistently \cite{ Rockwell2001CultureGA, Dress2008RegionalVI, oprea-magdy-2019-exploring}. 

In light of this, recent works in the multimodal domain have explored techniques like prompt‑learning and adapter‑learning with pretrained language models (PLMs) for few‑shot sarcasm \cite{Liang2022ModularAP, Jana2024ContinuousAM} and sentiment analysis \cite{Yu2022FewShotMS, Yu2022UnifiedMP, Yang2022FewshotMS, Wu2024MixtureofPromptExpertsFM}. These techniques have shown promise in few-shot settings, allowing the model to adapt effectively by introducing only a small number of learnable parameters while keeping the large pretrained backbone intact. \textit{The core motivation behind adopting these techniques is to achieve parameter-efficient learning, as large parameter size tend to overfit when training data is scarce.} However, while these methods mitigate overfitting, they often underperform because they have limited data to learn and generalize from—a challenge we refer to as \textbf{supervision scarcity}. From Table \ref{tab:kd_ablation}, we observe that across both datasets, vanilla PEFT methods (without distillation) fall short of their distillation-enhanced variants by approximately 1.7–3.5\% in accuracy under the 1\% data setting.

To mitigate this supervision scarcity problem, we take inspiration from the teacher-student framework for \textit{knowledge distillation (\textbf{KD})} \cite{Hinton2015DistillingTK} and propose \textbf{PEKD} (Parameter‑Efficient Knowledge Distillation), a plug‑and‑play framework designed for few‑shot multimodal sarcasm detection. This framework allows any model built with parameter‑efficient modules, such as LoRA \cite{Hu2021LoRALA}, adapter \cite{Houlsby2019ParameterEfficientTL}, or prompt tuning \cite{Lester2021ThePO}, to be easily plugged in as the student, while an expert model trained on large-scale sarcasm data serves as the teacher. This setup enables the student to generalize from limited examples without overfitting, while learning inductive biases from the teacher through knowledge distillation. A major challenge in distillation is unreliable teacher predictions, which can mislead the student. We address this with an \textit{entropy-aware gating mechanism} that weights the distillation loss based on teacher confidence, providing strong guidance when the teacher is confident and reducing its influence when the teacher is uncertain. This ensures effective knowledge transfer while avoiding noisy signals. 

Compared to SOTA multimodal baselines, our framework PEKD helps PEFT methods achieve superior performance in the 1\% data regime (\textbf{$\Delta$2.2\% to $\Delta$5.2\%}). Additionally, when compared against SOTA LVLMs, PEKD consistently helps PEFT techniques outperform across 5/10/20 shot settings, with LoRA even outperforming the best LVLM LLaMA-3.2-11B using 18× fewer trainable parameters. In addition to strong quantitative results, we conduct qualitative evaluations, including embedding space visualization, student-teacher representation alignment, and error mitigation analysis to understand how distillation benefits different PEFT variants. \textbf{Our contributions are:}
\begin{enumerate}
    \item We propose PEKD, the first teacher-student framework combining PEFT with entropy-aware knowledge distillation for few-shot multimodal sarcasm detection.
    
    \item We systematically evaluate and compare three PEFT variants—LoRA, Adapters, and Prompt Tuning, under our KD framework, and show that distillation consistently enhances their performance.
    
    \item We conduct a comprehensive empirical analysis to reveal how KD improves representation alignment, prediction confidence, and reduces mismatched errors between student and teacher.
\end{enumerate}

\section{Related Work}
\noindent\textbf{Image-Text Sarcasm Detection}\\
The rise of visual content on social media has spurred interest in multimodal sarcasm detection. Early efforts by \citet{Schifanella2016DetectingSI} used hand-crafted image-text features, followed by hierarchical and contrastive fusion approaches \cite{cai-etal-2019-multi, xu-etal-2020-reasoning, pan-etal-2020-modeling} that model intra-modal and inter-modal incongruity. Graph-based techniques \cite{Liang2021MultiModalSD, Liang2022MultiModalSD} model token-level and global-level incongruity while knowledge-enriched \cite{liu-etal-2022-towards-multi-modal} models enhanced cross-modal reasoning. More recent advances include dynamic routing on modalities for capturing dominant modalitiy of sarcasm. \cite{tian-etal-2023-dynamic}, \citet{qin-etal-2023-mmsd2} improved MMSD dataset, and proposed CLIP-based modeling.\citet{Tang2024LeveragingGL} used retrieval-augmented instruction tuning while \citet{10.1145/3664647.3680914} introduced parameter-efficient learning using mixture of adapters. \citet{Jana2024ContinuousAM} proposed the first dedicated few-shot multimodal sarcasm detection model using prompt-tuning approach on BERT.

In this study, we address the problem of few-shot multimodal saracsm detection with PEFT and distillation techniques.

\noindent\textbf{Multimodal Few-Shot Detection}\\
Early efforts for few-shot sentiment analysis used prompting and prompt tuning in PLM \cite{Yu2022FewShotMS}. \citet{Yu2022UnifiedMP} used a pre-training task to align image prompts before downstream sentiment analysis task. \citet{Yang2022FewshotMS} fused discrete prompts through bayesian fusion for improving sentiment detection. For object detection \cite{Zhou2021LearningTP} inserted prompts within CLIP text encoders while \cite{Zhou2022ConditionalPL} used image-conditioned prompts. \citet{Gao2021CLIPAdapterBV} introduced adapters in CLIP for object detection. \citet{Jana2024ContinuousAM} used prompt tuning with attentive prompts for few-shot multimodal sarcasm detection.

Our approach differs orthogonally by leveraging knowledge distillation with PEFT methods under a teacher-student setup for robust few-shot multimodal sarcasm detection.

\section{Task Definition}

The few‑shot multimodal sarcasm detection task is a binary classification problem: given an image–text pair $\mathcal{X} = (I,T)$, the goal is to predict $y \in {0,1}$, where $1$ denotes sarcasm. In the few‑shot setting, we have a small support set $\mathcal{S} = \{ (\mathcal{X}_i, y_i) \}_{i=1}^{N}$ with $N$ equally split between sarcastic and non‑sarcastic classes.

\begin{figure*}[ht]
    \centering
    \includegraphics[width=0.9\textwidth]{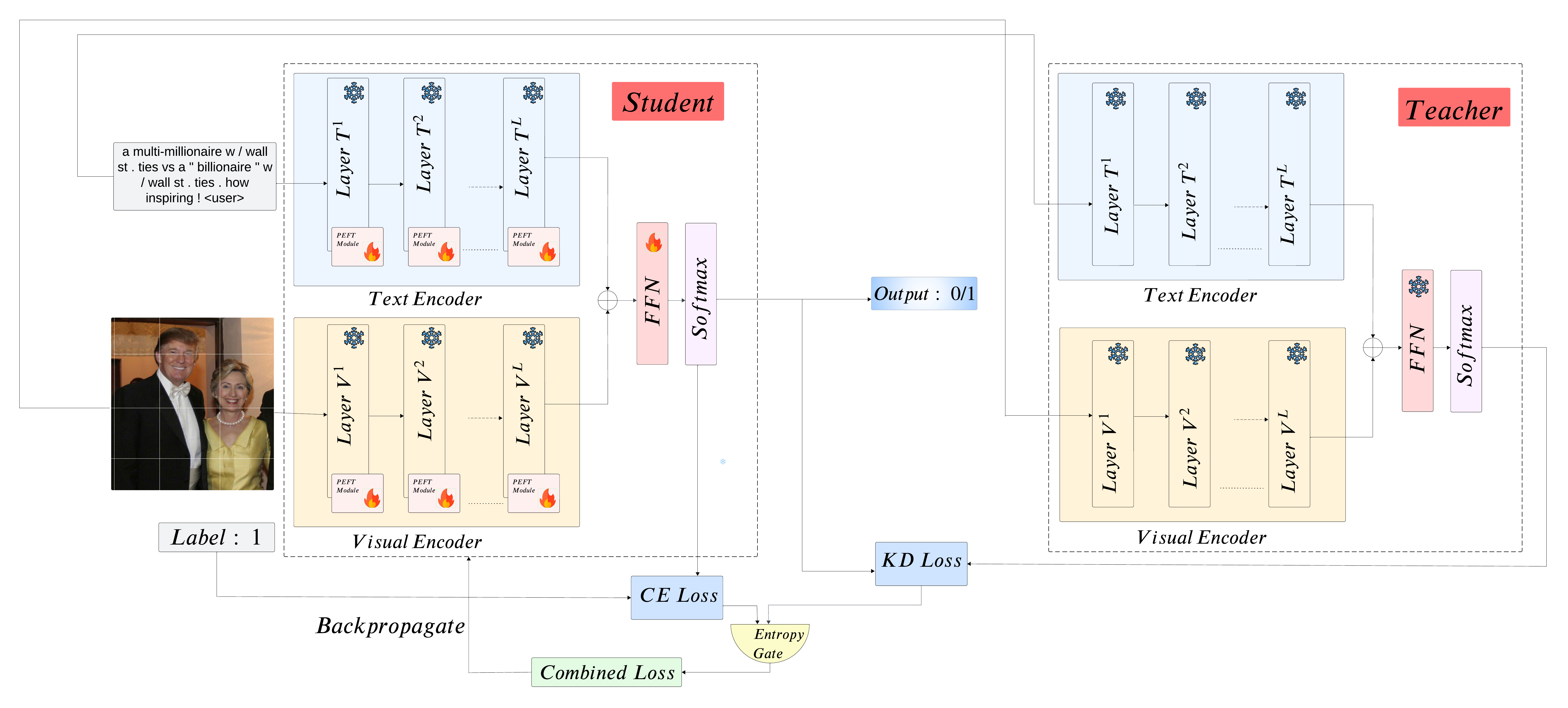} 
    \caption{\label{fig:pekd}Architecture of PEKD framework.}
\end{figure*}

\section{Proposed Approach}
We choose CLIP \cite{Radford2021LearningTV} as the backbone for both student and teacher due to its strong multimodal grounding and proven effectiveness on sarcasm detection tasks \cite{qin-etal-2023-mmsd2, 10.1145/3664647.3680914}, however, it can be extended to other vision-language models as well.

\subsection{CLIP Preliminaries}
\textbf{Vision Encoder:}
The input image \( I \) is divided into patches and passed through
\( L \) transformer blocks:
\begin{equation}
[\boldsymbol{z^i}, E^i] = \mathcal{E}_v^i([\boldsymbol{z^{i-1}}, E^{i-1}]),
\quad i = 1, \dots, L
\end{equation}
where \(\mathcal{E}_v^i\) is the visual transformer at layer \(i\), \( E^i \in \mathbb{R}^{m \times d_v} \) denotes the patch embeddings from the \( i^\text{th} \) layer, and \( \boldsymbol{z^i} \in \mathbb{R}^{1 \times d_v} \) is the embedding of the learnable class token.
The final visual representation \(\bold{h_{img}} \) is obtained by projecting the output class token from the last layer $L$:
\vspace{-1.5mm}
\begin{equation}
\bold{h_{img}} = \text{Proj}_{\mathcal{E}_v}(\boldsymbol{z^L}),
\quad i \in \mathbb{R}^{d}
\vspace{-2mm}
\end{equation}

\textbf{Text Encoder:} A given input text sequence \( l \) is tokenized, embedded and passed through the \( L \) text transformer blocks \( \mathcal{E}_t \):
\vspace{-1.5mm}
\begin{equation}
W^i = \mathcal{E}_t^i(W^{i-1}),
\quad i = 1, \dots, L
\vspace{-2mm}
\end{equation}
where \( W^i \in \mathbb{R}^{n \times d_t} \) represents the text embeddings from layer \( T^{i} \).
To derive the final textual representation, the embedding of the last token in the final layer \(L\) is projected into the shared embedding space:
\vspace{-1.5mm}
\begin{equation}
\bold{h_{txt}} = \text{Proj}_{\mathcal{E}_t}(W^L[-1]),
\quad t \in \mathbb{R}^{d}
\vspace{-0.5mm}
\end{equation}



\subsection{Methodology}
We propose PEKD, a parameter-efficient framework for few-shot multimodal sarcasm detection, illustrated in Fig.~\ref{fig:pekd}. It consists of two components: (a) A teacher model fully fine-tuned (updating all its parameters) on a large-scale sarcasm dataset, (b) a student augmented with PEFT modules for efficient adaptation. During training, both teacher and student process the same input, and their outputs are used to compute the KD loss along with the task loss. An entropy-aware gating mechanism combines these losses to regulate the teacher’s influence based on confidence, and only the student’s PEFT modules are updated. We elaborate on the details of the teacher, the student, and the fine-tuning of the student in the subsections below.

\subsection{Teacher Model}
Let $\mathcal{T}$ denotes the teacher, a pretrained CLIP model fine-tuned on a large sarcasm dataset. 
Due to its extensive parameterization and access to abundant training data, $\mathcal{T}$ captures intricate, cross-modal patterns of sarcasm. 
In few-shot settings, where direct fine-tuning of a large model can lead to overfitting, the teacher provides strong supervision and guides a parameter-efficient student, facilitating the transfer of its rich, sarcasm-sensitive knowledge. The teacher predicts the sarcasm label as:
\vspace{-1.5mm}
\begin{align}
    \boldsymbol{y_\mathcal{T}} &= \mathrm{Softmax}(W_\mathcal{T} \cdot (\bold{h_{img}} \oplus \bold{h_{txt}}))~
    \vspace{-2mm}
\end{align}

where $\bold{h_{img}}$ and $\bold{h_{txt}}$ are the image and text embeddings obtained from the respective CLIP encoders, $W_\mathcal{T}$ is the projection layer, $\oplus$ is the concatenation operator and $\boldsymbol{y_\mathcal{T}}$ is the soft logits from the teacher.

\subsection{Parameter-Efficient Student Model}
We design the student $\mathcal{S}$ as a parameter-efficient model that can be adapted to few-shot settings using a range of PEFT techniques. In this work, we focus on three such techniques, namely, \emph{adapters}, \emph{prompt-tuning}, and \emph{LoRA} and apply them to CLIP. Our framework is flexible, allowing any PEFT method to be plugged in as the student model. 
Each technique adds only a small number of learnable parameters, keeping the rest of the CLIP backbone frozen, which reduces overfitting in few‑shot settings. The trainable parameter sizes are specified in Table \ref{tab:param_sizes}. We outline the adaptations of these techniques in CLIP in the following subsections.

\subsubsection{Adapter-CLIP}
In this setup, we insert \emph{adapters} into every layer of CLIP’s text and visual encoders. These adapters are lightweight bottleneck layers, consisting of a down‑projection, a non‑linearity, and an up‑projection layer.
Adapter $Ad$ is realized as:
\vspace{-1.5mm}
\begin{equation}
 Ad(\boldsymbol{h}) = M_{up}\cdot\theta(M_{down}\cdot \boldsymbol{h})   
 \vspace{-2mm}
\end{equation}
where $M_{up}$ and $M_{down}$ are upsample and downsample projection layers respectively and $\theta$ is ReLU non-linearity. The visual and text encoders in equations (1) and (3) can be modified as: 
\vspace{-1.5mm}
\begin{align}
[\boldsymbol{z^{i}}, E^{i}] &= \mathcal{E}_v^{i}([\boldsymbol{z^{i-1}}, E^{i-1}]) \notag \\
&\quad + \text{Ad}_{\mathcal{E}_v}^{i}([\boldsymbol{z^{i-1}}, E^{i-1}])\notag \\
& \hspace{2cm} i = 1,.., L \\
[W^{i}] &= \mathcal{E}_t^{i}([W^{i-1}]) + Ad_{\mathcal{E}_t}^{i}([W^{i-1}]) \notag \\
& \hspace{2cm} i = 1,.., L 
\vspace{-2mm}
\end{align}

where $Ad_{\mathcal{E}_v}^{i}$ and $Ad_{\mathcal{E}_t}^{i}$ are the adapter layers in the $i^{th}$ transformer block of the visual and text encoders, respectively.

The insertion of adapters reduces the model's trainable parameter size to around $3\sim4\%$ of the original CLIP model. During fine-tuning, only the adapters get trained while the CLIP backbone is frozen. 

\subsubsection{Prompt-CLIP}
In this configuration, we introduce a small set of learnable embeddings, called \emph{prompts}, that are added to the input of the text and visual encoders at every layer. After applying prompts, the visual and text encoders in equations (1) and (3)  can be reformulated as:
\vspace{-1.5mm}
\begin{align}
[\boldsymbol{z^{i}}, E^{i}, P_{\mathcal{E}_v}^{i}] &= \mathcal{E}_v^{i}([\boldsymbol{z^{i-1}}, E^{i-1}, P_{\mathcal{E}_v}^{i-1}]) \notag \\
& \hspace{2cm} i = 1,.., L \\
[P_{\mathcal{E}_t}^{i}, W^{i}] &= \mathcal{E}_t^{i}([P_{\mathcal{E}_t}^{i-1}, W^{i-1}]) \notag \\ 
& \hspace{2cm} i = 1,.., L 
\vspace{-2.5mm}
\end{align}

where $P_{\mathcal{E}_v}^{i-1}$ and $P_{\mathcal{E}_t}^{i-1}$ are learnable prompt embeddings added to the input of the visual and text encoders $i$. This approach reduces the number of trainable parameters to roughly $0.02\sim0.03\%$ of the original CLIP model. During fine-tuning, only the prompts are trained, while the CLIP backbone remains frozen.

\subsubsection{LoRA-CLIP}
LoRA (Low-Rank Adaptation) updates a pre-trained weight matrix $W \in \mathbb{R}^{d_1 \times d_2}$ using a low-rank decomposition $\Delta W = B A$, where $A \in \mathbb{R}^{r \times d_2}$, $B \in \mathbb{R}^{d_1 \times r}$, and $r \ll \min(d_1, d_2)$. For input $X$, the adapted projection becomes:
\begin{align}
\boldsymbol{h} = W X + \gamma (BA) X,
\end{align}
with $\gamma$ as a scaling factor. We apply LoRA to the query, key, and value matrices of each attention layer in both text and vision encoders:
\begin{align}
  Q_{Lo} &= XW_q + \gamma(B_q A_q)X \\
  K_{Lo} &= XW_k + \gamma(B_k A_k)X \\
  V_{Lo} &= XW_v + \gamma(B_v A_v)X
\end{align}
The attention operation now becomes:
\begin{align}
\mathrm{Attn} &= \mathrm{Softmax}\!\left(\frac{Q_{Lo} K_{Lo}^T}{\sqrt{d_k}}\right) V_{Lo}
\end{align}
LoRA reduces trainable parameters by around 1.9\% of the original CLIP. Only the low-rank matrices $A$ and $B$ are optimized during training.

\begingroup
\renewcommand{\arraystretch}{1.3} 
\begin{table}[t]
\centering
\scriptsize
\adjustbox{max width=0.6\columnwidth}{
\begin{tabular}{lc}
\toprule
\textbf{Model} & \textbf{Trainable Parameters} \\
\midrule
CLIP ViT-B/16 (Teacher) & 149 M \\
LoRA-CLIP (Student) & 2.9 M \\
Prompt-CLIP (Student) & 0.03 M \\
Adapter-CLIP (Student) & 4.1 M \\
\bottomrule
\end{tabular}
}
\caption{Trainable parameter sizes of teacher and student models.}
\label{tab:param_sizes}
\end{table}
\endgroup

\begingroup
\begin{table}[t]
\centering
\renewcommand{\arraystretch}{1.25}
\tiny
\setlength{\tabcolsep}{4pt}
\begin{tabular}{lccc}
\toprule
\textbf{Model} & \textbf{Train} & \textbf{Valid} & \textbf{Test} \\
\midrule

\rowcolor{gray!15}
\multicolumn{4}{c}{\textbf{MMSD}} \\
Teacher (99\%) & 8543 / 11075 / 19618 & 860 / 1351 / 2211 & 959 / 1450 / 2409 \\
Student (1\%)   & 99 / 99 / 198         & 99 / 99 / 198     & 959 / 1450 / 2409 \\
\midrule

\rowcolor{gray!15}
\multicolumn{4}{c}{\textbf{MMSD2.0}} \\
Teacher (99\%) & 9477 / 10141 / 19618  & 943 / 1269 / 2212 & 1037 / 1072 / 2409 \\
Student (1\%)   & 99 / 99 / 198         & 99 / 99 / 198      & 1037 / 1072 / 2409 \\

\bottomrule
\end{tabular}
\caption{\label{tab:dataset_stats}
Train, valid, and test splits (Pos/Neg/Total) for teacher and student. Student sees only 1\% of the training data, while the teacher sees the remaining 99\%.}
\end{table}
\endgroup

\subsection{Fine-tuning the Student Model}
To fine-tune the student, we use knowledge distillation (KD) in addition to the task-specific loss to enable the student to learn from both ground truth labels and the teacher’s rich output distribution. This allows the student to mimick the teacher’s rich sarcasm-specific cross-modal representations, which is difficult for the student to generalize from limited training examples. 

We follow the same operations as the teacher (similar to equation 5) to get the final soft logits $ \boldsymbol{y_\mathcal{S}}$ from the student. We fine-tune $\mathcal{S}$ on the few-shot data split, updating only the PEFT modules and the projection layer $W_{\mathcal{S}}$. The combined objective for fine-tuning the student is:
\vspace{-2mm}
\begin{align}
    \mathcal{L} = g \mathcal{L}_{CE} + (1-g) \mathcal{L}_{KD}
\vspace{-3.5mm}
\end{align}
where, $\mathcal{L}_{CE}$ is the task-specific cross-entropy loss while $\mathcal{L}_{KD}$ is the knowledge-distillation loss from the teacher to the student.

The KD loss is computed as the KL divergence between teacher and student predictions:  
\begin{align}
\mathcal{L}_{KD} = T^2 \sum_{i=1}^{C} y_\mathcal{T}(i) \log \frac{y_\mathcal{T}(i)}{y_\mathcal{S}(i)},
\end{align}
where \( C \) is the number of classes and \( T^2 \) compensates for the temperature scaling effect. There might be cases when the teacher's predictions are less confident. In such scenarios, we want the student to rely on learning from the data rather than rely on the teacher. To achieve this, we introduce an entropy-aware gating parameter \( g \) based on normalized entropy of the teacher:  
\begin{align}
g = \frac{-\sum_{i=1}^{C} y_\mathcal{T}(i) \log y_\mathcal{T}(i)}{\log C},
\end{align}
where \(\log C\) is the maximum possible entropy, ensuring \( g \in [0,1] \).  

\textbf{Lemma.} Weighting \(\mathcal{L}_{KD}\) by \( (1-g) \) ensures it approaches zero as teacher uncertainty increases (\(g \to 1\)) and remains intact when the teacher is confident (\(g \to 0\)). (Proof in Appendix~\ref{app:proof}.)

\textit{Alternative Variant.} We also experimented with a variant that sets the KD loss to zero when the teacher’s prediction is incorrect; however, this approach underperformed compared to entropy-based gating. Detailed discussion is provided in Appendix~\ref{app:hard_gate}.

\section{Experiments}

\subsection{Datasets}

We assess our approach, PEKD, on the few‑shot splits of MMSD \cite{cai-etal-2019-multi} and MMSD2.0 \cite{qin-etal-2023-mmsd2}, proposed in the study \cite{Jana2024ContinuousAM}. They extract two 1\% few‑shot splits for each dataset, with an equal number of samples per class. The size of the test set remains unchanged from the original dataset. \textbf{We utilize the 1\% split to train the student while the remaining 99\% to train the teacher}. Detailed statistics are provided in Table \ref{tab:dataset_stats}.

\subsection{Experimental Settings}

We employ \texttt{ViT-B/16 CLIP} backbone for both the teacher and student. 
Few‑shot training can exhibit significant performance variability. To capture this, we train the student 6 times (3 trials × 2 splits) for each dataset and report the average Accuracy (Acc), average Macro‑F1 (F1), and the standard deviation across all six runs. Additional hyperparameter details for training the teacher and the student are reported in Appendix \ref{app:hyperparameter}.

\subsection{Baselines}



We evaluate our approach against two groups of baselines: PEFT and non-PEFT-based. For the non-PEFT based multimodal baselines, we consider the following  SOTA models for sarcasm detection; \textbf{HFM} \cite{cai-etal-2019-multi}, \textbf{Att-BERT} \cite{pan-etal-2020-modeling}, \textbf{HKE} \cite{liu-etal-2022-towards-multi-modal}, \textbf{DIP} \cite{Wen2023DIPDI}, \textbf{MV-CLIP} \cite{qin-etal-2023-mmsd2}, \textbf{DynRT} \cite{tian-etal-2023-dynamic} and \textbf{RAG-LLaVA} \cite{Tang2024LeveragingGL}. For the PEFT-based multimodal group, we consider \textbf{PVLM} \cite{Yu2022FewShotMS}, \textbf{UP-MPF} \cite{Yu2022UnifiedMP}, and \textbf{CAMP} \cite{Jana2024ContinuousAM} which are based on prompt-tuning in PLMs.  \textbf{CoOp} \cite{Zhou2021LearningTP} and \textbf{CoCoOp} \cite{Zhou2022ConditionalPL} are prompt-tuning adaptations for CLIP. \textbf{MoBA} \cite{10.1145/3664647.3680914} is an adapter-based adaptation of CLIP. 
The details of these baselines are provided in the Appendix \ref{app:baselines}. We evaluate all baselines on the few-shot data splits and report the performances.

\begin{table}[t!]
\centering
\renewcommand{\arraystretch}{1.5}
\setlength{\tabcolsep}{7pt} 
\begin{adjustbox}{max width=0.95\columnwidth}
\begin{tabular}{lllll}
\toprule
& \multicolumn{2}{c}{\textbf{MMSD}} & \multicolumn{2}{c}{\textbf{MMSD2.0}} \\
\cmidrule(lr){2-3} \cmidrule(lr){4-5}
\textbf{Method} & \textbf{Acc} & \textbf{F1} & \textbf{Acc} & \textbf{F1} \\

\midrule

\multicolumn{5}{c}{\cellcolor{gray!10}\textbf{Multimodal (Non PEFT)}} \\
\midrule
HFM & 0.612 (1.3) & 0.598 (1.1) & 0.561 (0.2) & 0.361 (0.3) \\
Att-BERT & 0.707 (1.7) & 0.696 (1.3) & 0.659 (1.6) & 0.683 (1.8) \\
HKE & 0.503 (2.3) & 0.667 (2.8) & 0.408 (1.5) & 0.579 (1.3) \\
DIP & 0.704 (2.7) & 0.698 (2.3) & 0.685 (2.8) & 0.658 (2.6) \\
DynRT & 0.583 (0.1) & 0.487 (0.6) & 0.518 (2.9) & 0.513 (3.2) \\
MV-CLIP & 0.780 (0.2) & 0.770 (0.2) & 0.742 (0.4) & 0.740 (0.3) \\
RAG-LLaVA & 0.483 (9.1) & 0.406 (4.6) & 0.569 (0.1) & 0.446 (8.3) \\
Teacher (CLIP) & 0.777 (0.5) & 0.768 (0.6) & 0.740 (1.2) & 0.739 (1.1) \\

\midrule
\multicolumn{5}{c}{\cellcolor{gray!10}\textbf{Multimodal (PEFT)}} \\
\midrule
PVLM & 0.712 (0.6) & 0.699 (0.2) & 0.665 (2.2) & 0.658 (2.1) \\
UP-MPF & 0.707 (2.4) & 0.701 (2.6) & 0.669 (0.4) & 0.663 (0.1) \\
CAMP & 0.729 (0.9) & 0.717 (1.0) & 0.692 (2.8) & 0.681 (2.3) \\
CoOp & 0.772 (1.6) & 0.769 (1.5) & 0.759 (0.9) & 0.759 (0.8) \\
CoCoOp & 0.782 (1.0) & 0.779 (1.0) & 0.746 (1.6) & 0.745 (1.6) \\
MoBA & 0.799 (1.2) & 0.790 (1.1) & 0.758 (0.7) & 0.753 (0.9) \\

\midrule
\multicolumn{5}{c}{\cellcolor{gray!10}\textbf{PEKD (Ours)}} \\
\midrule

\rowcolor{top2}
Adapter-CLIP (w/ KD) & 0.824 (0.2) & 0.819 (0.3) & 0.799 (0.1) & 0.792 (0.3) \\
\rowcolor{top3}
Prompt-CLIP (w/ KD) & 0.821 (0.3) & 0.811 (0.4) & 0.781 (0.5) & 0.798 (0.4)\\
\rowcolor{top1}
LoRA-CLIP (w/ KD) & 0.845 (0.3) & 0.843 (0.1) & 0.811 (0.2) & 0.811 (0.2) \\


\bottomrule
\end{tabular}
\end{adjustbox}
\caption{\label{tab:main_results}
Performance comparison of multimodal methods across few-shot MMSD and MMSD2.0 datasets. Our PEKD variants achieves SOTA performance. Numbers in brackets indicate standard deviation. The 
\colorboxlegend{top1}{best}, 
\colorboxlegend{top2}{second-best}, and 
\colorboxlegend{top3}{third-best} results are highlighted respectively. Our method outperforms baselines significantly with p < 0.05.
}
\vspace{-3mm}
\end{table}

\section{Main Results}
Following \citet{Jana2024ContinuousAM}, we report the few-shot performance of the models in Table \ref{tab:main_results}.  Our observations are: \textbf{(1)} PEFT methods often outperform or match their non-PEFT counterparts by leveraging task-specific parameters while keeping the pretrained backbone frozen, thus reducing overfitting. \textbf{(2)} When we train the teacher, \texttt{Teacher (CLIP)} (row 8 in Non-PEFT), on 1\% few-shot split, we observe that it overfits and performs poorly due to its large parameter size. In contrast, the student models (\texttt{LoRA-CLIP (w/ KD)}, \texttt{Adapter-CLIP (w/ KD)}, and \texttt{Prompt-CLIP (w/ KD)}) benefit from KD by acquiring inductive biases from the teacher (trained on the remaining 99\% split). As a result, they outperform the best baseline on MMSD, \texttt{MoBA} by gains ranging from \textbf{$\Delta$2.2\% to $\Delta$4.6\%} in Acc and \texttt{CoOp} on MMSD2.0 by \textbf{$\Delta$2.2\% to $\Delta$5.2\%}. \textbf{(3)} Among the students, \texttt{LoRA-CLIP (w KD)} demonstrates superior performance in the KD setup due to LoRA's ability to directly modify the attention mechanism of the pretrained model. By injecting low-rank updates into the $Q, K, V$ matrices of the attention layers, LoRA enables the student to more precisely align its internal attention patterns with those of the teacher, unlike prompts (which affect only inputs) or adapters (which act more peripherally).

\section{Comparison with LVLMs}
\begin{figure}[t!]
    \centering
    \includegraphics[width=0.45\textwidth, height=0.27\textwidth]{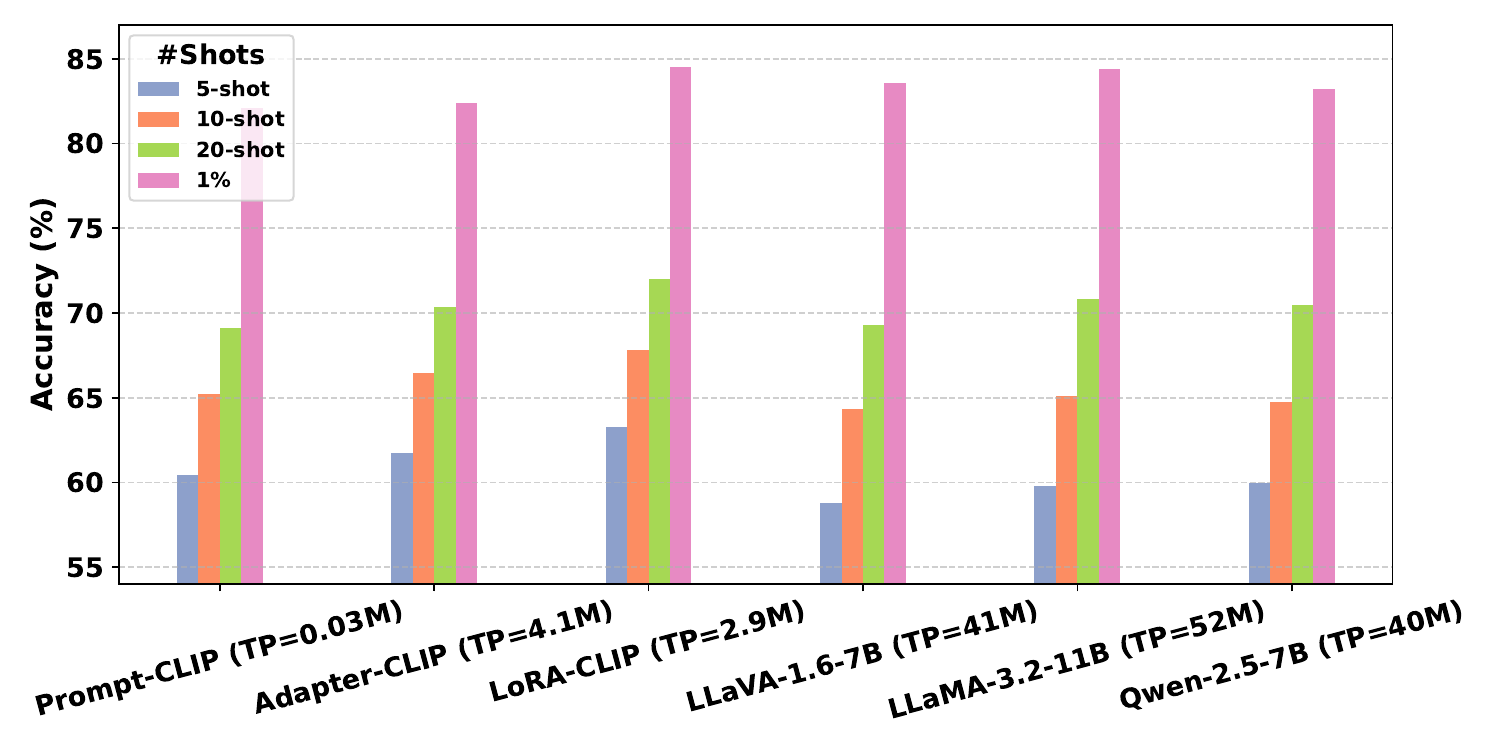} 
    \caption{\label{fig:comp_with_lvlms} Comparison of PEKD-based models with LVLMs in 5/10/20 shots and 1\% data setting for MMSD2.0 dataset. TP denotes trainable parameters.}
\end{figure}

We compare our PEKD-based models with SOTA LVLMs, LLaVA-1.6-7B \cite{Liu2023ImprovedBW}, LLaMA-3.2-11B \cite{Dubey2024TheL3}, and Qwen-2.5-VL-7B \cite{Bai2025Qwen25VLTR} across extremely low-resource settings: 5/10/20-shots, as well as 1\% supervision, as shown in Fig \ref{fig:comp_with_lvlms} for MMSD2.0 dataset. For fine-tuning of these LVLMs, we resort to  LoRA with rank 8, due to resource constraints. Despite having drastically fewer trainable parameters (e.g., \texttt{Prompt-CLIP}: 0.03M, \texttt{Adapter-CLIP}: 4.1M, \texttt{LoRA-CLIP}: 2.9M vs. 40–52M for LVLMs), our models consistently outperform LVLMs in the 5/10/20-shot settings. Even at 1\% supervision, PEKD-based methods reach near-comparable accuracy to LVLMs while \texttt{LoRA-CLIP} outperforms the LVLMs in this setting.

\begingroup
\renewcommand{\arraystretch}{1.2}
\begin{table}[t]
\centering
\scriptsize
\setlength{\tabcolsep}{3pt}
\begin{tabular}{lcccccc}
\toprule
\multirow{2}{*}{\textbf{Model}} & \multicolumn{3}{c}{\textbf{MMSD}} & \multicolumn{3}{c}{\textbf{MMSD2.0}} \\
\cmidrule(lr){2-4} \cmidrule(lr){5-7}
 & Acc & F1 & $\Delta$ & Acc & F1 & $\Delta$ \\
\midrule
Prompt-CLIP (w/o KD) & 0.802 & 0.799 & \multirow{2}{*}{\textcolor{green!60!black}{\textbf{+1.9\%}}} & 0.762 & 0.762 & \multirow{2}{*}{\textcolor{green!60!black}{\textbf{+1.9\%}}} \\
Prompt-CLIP (w/ KD)  & 0.821 & 0.811 & & 0.781 & 0.798 & \\
\midrule
Adapter-CLIP (w/o KD) & 0.807 & 0.803 & \multirow{2}{*}{\textcolor{green!60!black}{\textbf{+1.7\%}}} & 0.771 & 0.778 & \multirow{2}{*}{\textcolor{green!60!black}{\textbf{+2.8\%}}} \\
Adapter-CLIP (w/ KD)  & 0.824 & 0.819 & & 0.799 & 0.792 & \\
\midrule
LoRA-CLIP (w/o KD) & 0.810 & 0.803 & \multirow{2}{*}{\textcolor{green!60!black}{\textbf{+3.5\%}}} & 0.783 & 0.782 & \multirow{2}{*}{\textcolor{green!60!black}{\textbf{+2.8\%}}} \\
LoRA-CLIP (w/ KD)  & 0.845 & 0.843 & & 0.811 & 0.811 & \\
\bottomrule
\end{tabular}
\caption{Ablation on KD. $\Delta$ shows \% improvement in Acc with KD. Green highlights indicate positive gains.}
\label{tab:kd_ablation}
\end{table}
\endgroup

\section{Ablation}
\subsection{KD Boosts PEFT Performance}
Table \ref{tab:kd_ablation} compares each PEFT variant with and without KD. Across all methods, KD consistently improves the performance of the student in the 1\% few-shot setting. In contrast, \texttt{LoRA-CLIP (w/o KD)}, \texttt{Adapter-CLIP (w/o KD)}, and \texttt{Prompt-CLIP (w/o KD)}, without KD, underperform due to insufficient supervision. The students trained with KD outperforms as we mitigate the supervision scarcity problem by inducting knowledge from the teacher.

\subsection{Effect of Entropy-Aware Gating}
\begin{figure}[t]
    \centering
    \includegraphics[width=0.4\textwidth, height=0.3\textwidth]{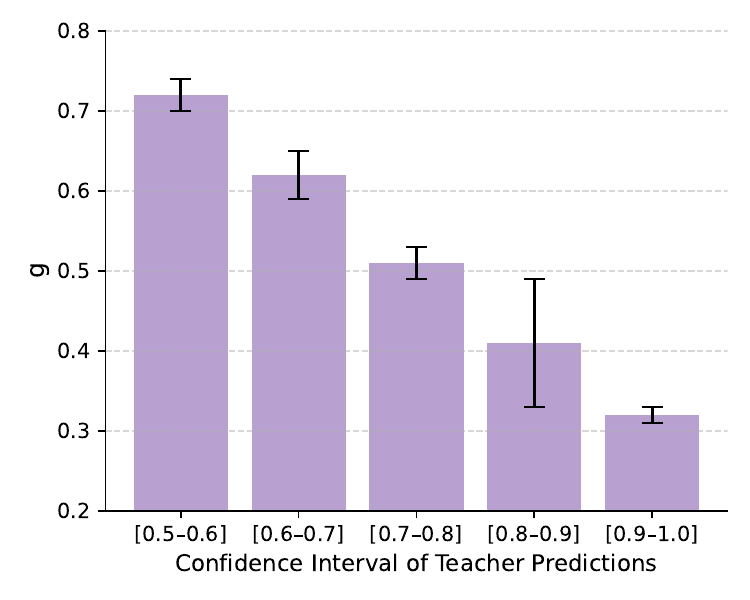} 
    \caption{\label{fig:entropy_vs_g}Effect of teacher confidence on KD gating: Lower confidence (higher uncertainty) results in reduced KD weight (1-g), demonstrating the adaptive behavior of entropy-aware gating.}
\end{figure}

To assess entropy-aware gating, we analyze its effect on the weighting of KD loss during training. Figure~\ref{fig:entropy_vs_g} shows that higher teacher confidence corresponds to lower entropy ($g$), resulting in larger weights $(1-g)$ for KD loss, thus amplifying reliable teacher signals. When confidence is low, $g$ increases, reducing KD influence. Table~\ref{tab:entropy_gating} confirms that entropy-based gating consistently improves performance across all student variants. 
\section{Analysis}

\begin{figure}[t!]
    \centering
    \includegraphics[width=0.45\textwidth]{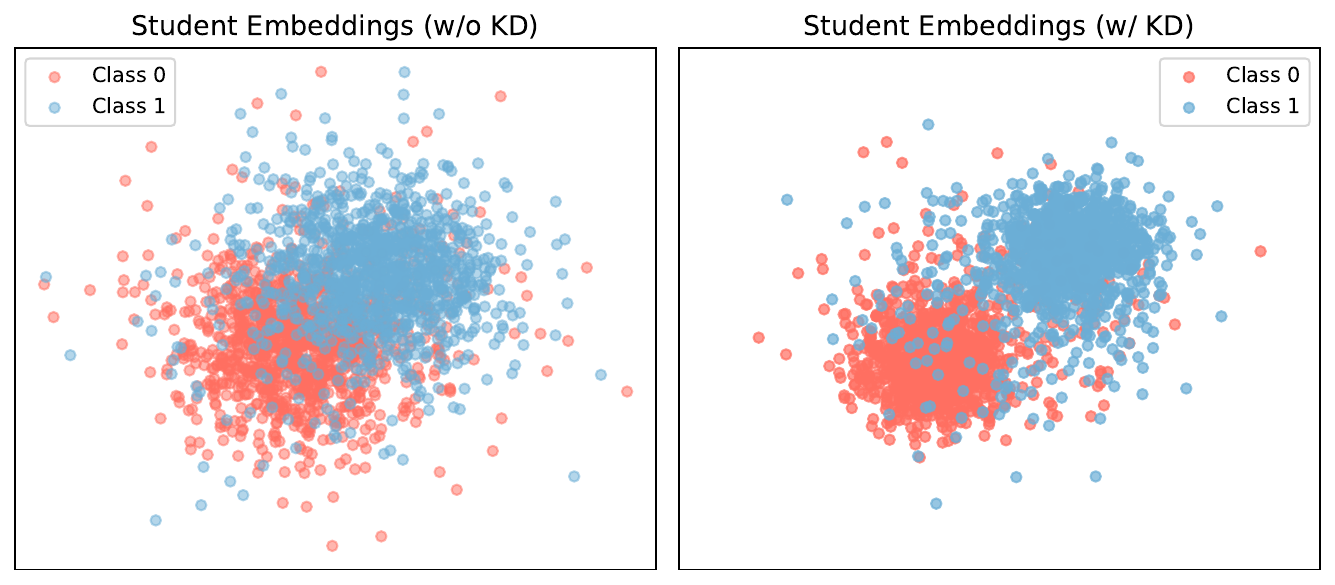} 
    \caption{\label{fig:logit_comparison}Comparison of student embeddings with and without KD.}
\end{figure}

\begin{figure}[t!]
    \centering
    \includegraphics[width=0.45\textwidth, height=0.2\textwidth]{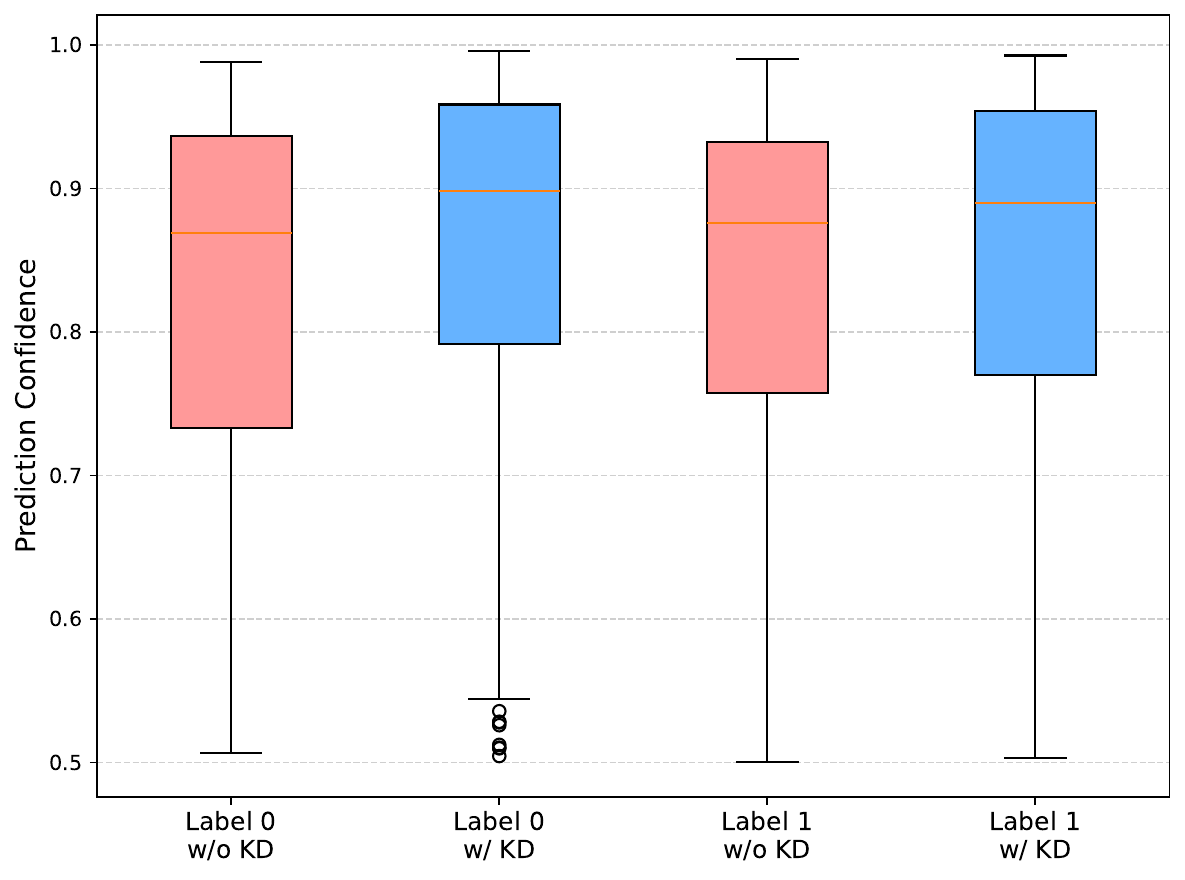} 
    \caption{\label{fig:confidence_comparison} Class-wise prediction confidence comparison of student with and without KD.}
\end{figure}

\subsection{Impact of KD on Student}
We study the impact of distilling knowledge from the teacher to the student from two perspectives: \textbf{(a)} \textit{Effect of KD on the student's embedding space:}We visualize the logit representations of the student model trained with and without KD in Fig \ref{fig:logit_comparison}. The KD-trained student exhibits better class-wise separation and structured embeddings, suggesting improved discriminative capacity and alignment with the teacher’s feature space. \textbf{(b)} \textit{Improved prediction confidence of the student:} As shown in Fig~\ref{fig:confidence_comparison}, the student model trained with KD exhibits consistently higher prediction confidence compared to its non-KD counterpart, across both class labels. This is evident from the upward shift in the median values and a tighter interquartile range.

\begingroup
\renewcommand{\arraystretch}{1.2}
\begin{table}[t]
\centering
\scriptsize
\setlength{\tabcolsep}{1pt}
\begin{tabular}{lcccccc}
\toprule
\multirow{2}{*}{\textbf{Model}} & \multicolumn{3}{c}{\textbf{MMSD}} & \multicolumn{3}{c}{\textbf{MMSD2.0}} \\
\cmidrule(lr){2-4} \cmidrule(lr){5-7}
 & w/o Gating & w/ Gating & $\Delta$ & w/o Gating & w/ Gating & $\Delta$ \\
\midrule
Prompt-CLIP  & 0.812 & \textbf{0.821} & \textcolor{green!60!black}{\textbf{+0.9\%}} & 0.771 & \textbf{0.781} & \textcolor{green!60!black}{\textbf{+1.0\%}} \\
Adapter-CLIP & 0.811 & \textbf{0.824} & \textcolor{green!60!black}{\textbf{+1.3\%}} & 0.777 & \textbf{0.799} & \textcolor{green!60!black}{\textbf{+2.2\%}} \\
LoRA-CLIP    & 0.835 & \textbf{0.845} & \textcolor{green!60!black}{\textbf{+1.0\%}} & 0.803 & \textbf{0.811} & \textcolor{green!60!black}{\textbf{+0.8\%}} \\
\bottomrule
\end{tabular}
\caption{\label{tab:entropy_gating}Ablation on Entropy-Aware Gating: Accuracy improvements when applying entropy-based gating for KD. $\Delta$ shows percentage gain.}

\end{table}
\endgroup

\begin{figure}[t]
    \centering
    \includegraphics[width=0.42\textwidth, height=0.38\textwidth]{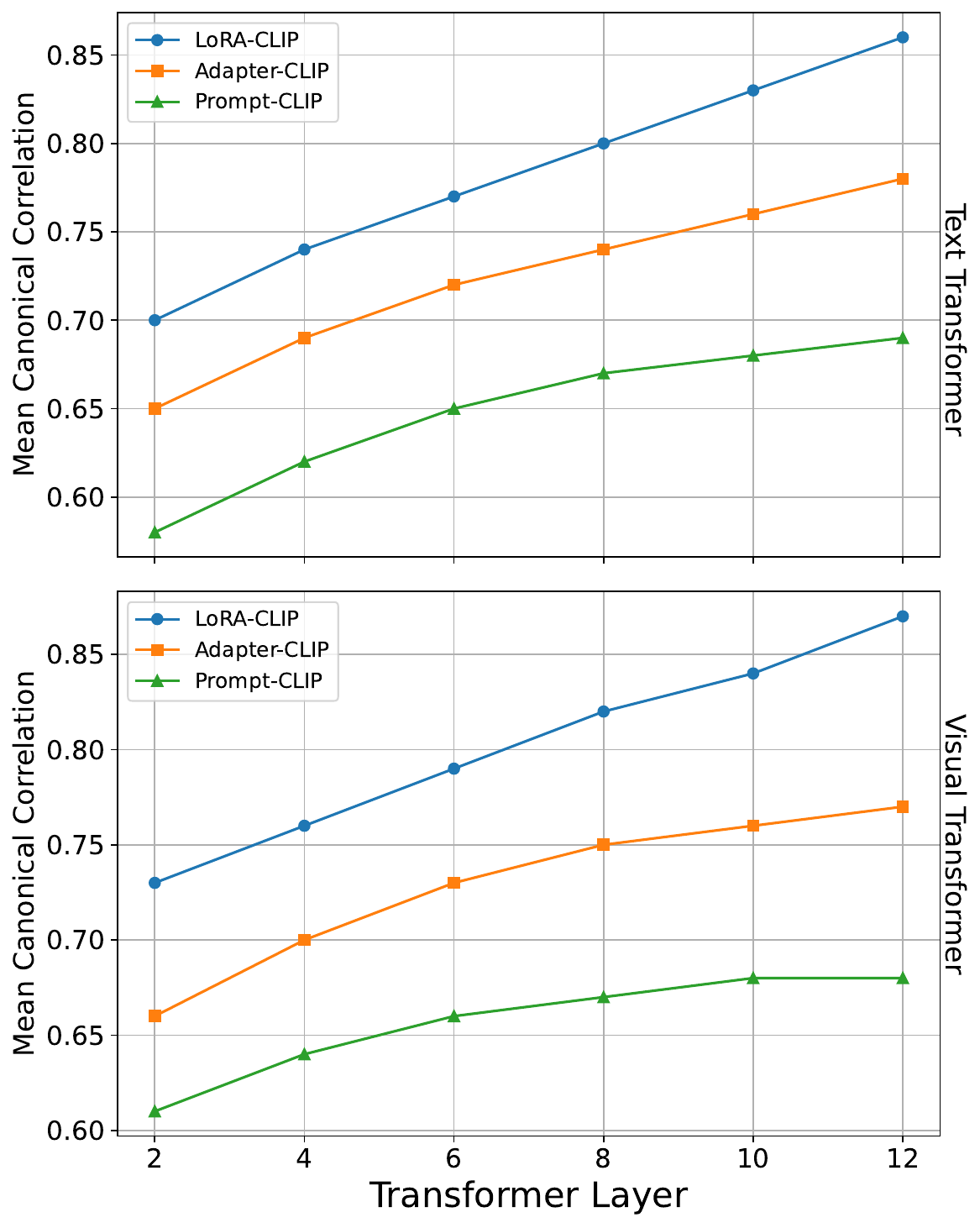} 
    \caption{\label{fig:layerwise_sim}Layer-wise CCA similarity between teacher and student models across text and visual transformers.}
\end{figure}

\subsection{Strength of LoRA-CLIP}
To evaluate the strength of \texttt{LoRA-CLIP} over \texttt{Adapter-CLIP} and \texttt{Prompt-CLIP}, we compute the the mean Canonical Correlation (CCA) between the hidden states of the teacher model and each student model across corresponding transformer layers, for both the text and vision branches. This metric quantifies how closely the student preserves the internal representational structure of the teacher. As shown in Fig~\ref{fig:layerwise_sim}, LoRA-CLIP consistently exhibits the highest CCA scores across all layers. This suggests that LoRA is more effective at preserving and transferring internal structural representations from the teacher model. Notably, the alignment gap between LoRA and other methods widens in deeper layers, indicating LoRA's superior ability to absorb hierarchical knowledge during distillation.

\subsection{Influence of KD on Student Errors}
\begin{figure}[t]
    \centering
    \includegraphics[width=0.4\textwidth, height=0.25\textwidth]{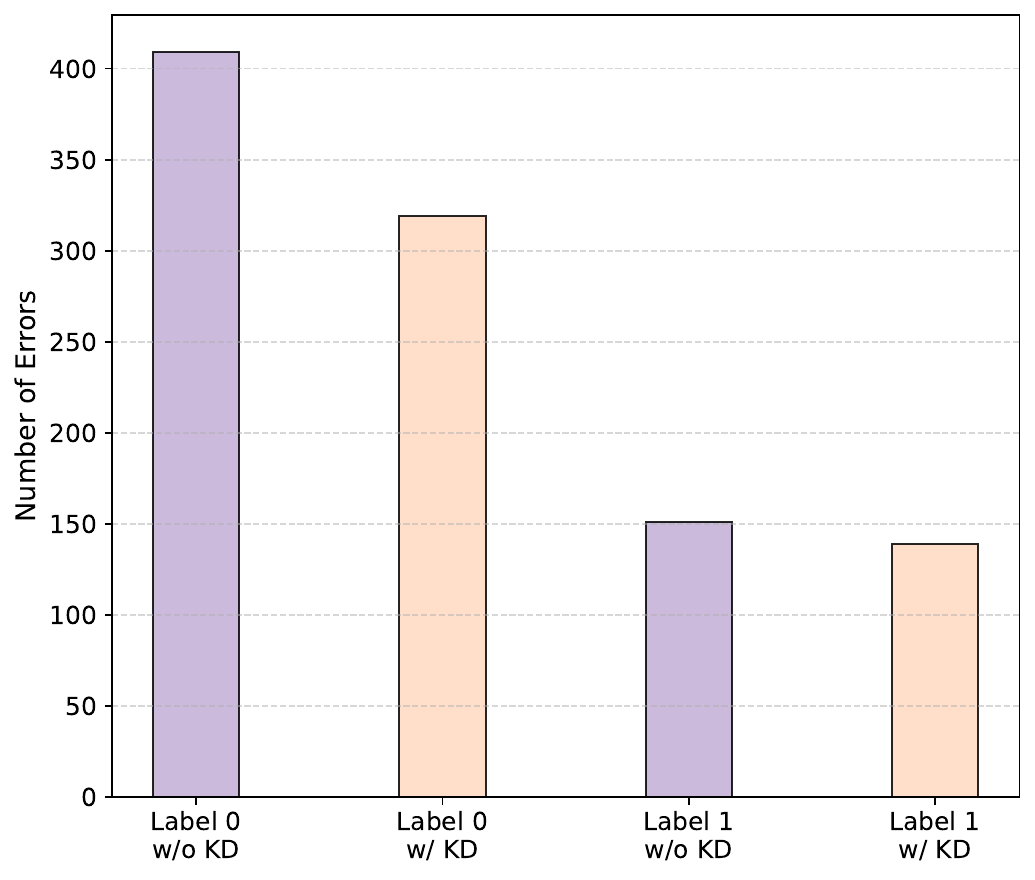} 
    \caption{\label{fig:error_comparison}Class-wise error comparison of the student with and without KD.}
\end{figure}

\begin{figure}[t]
    \centering
    \includegraphics[width=0.4\textwidth, height=0.25\textwidth]{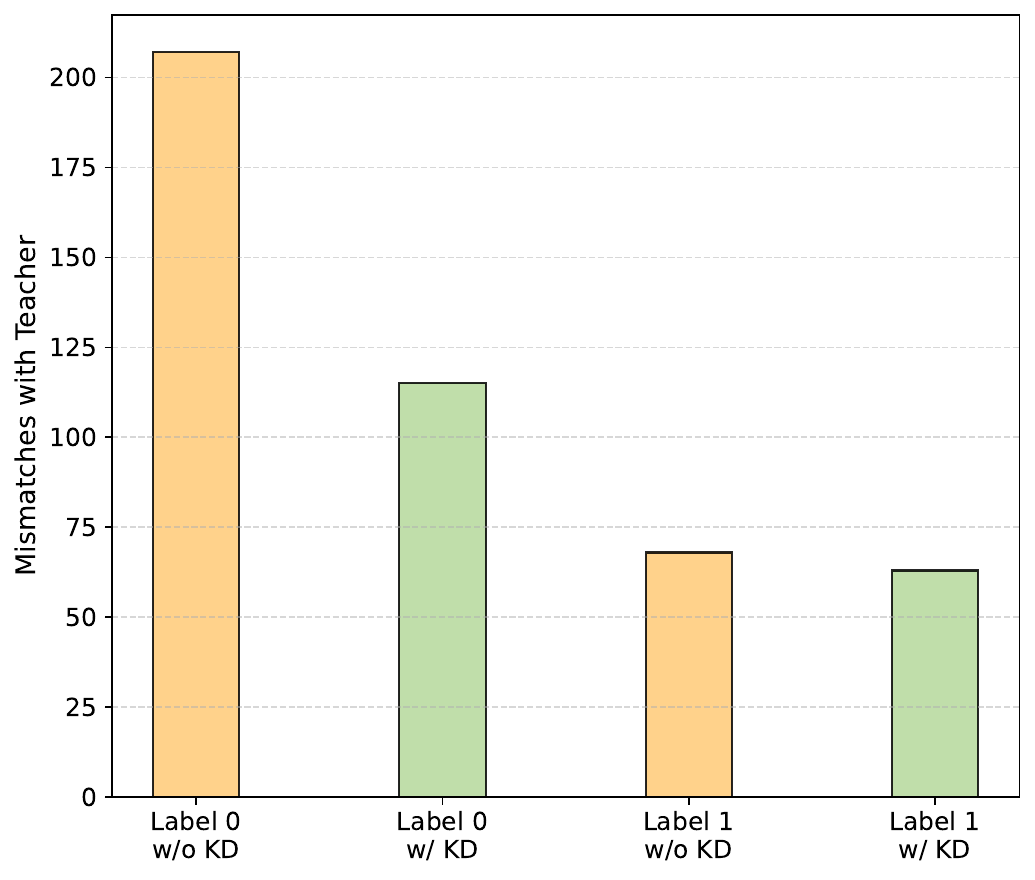} 
    \caption{\label{fig:mismatch_with_teacher}Unique prediction errors made by the student (i.e., errors made by the student but not by the teacher) across KD and non-KD settings.}
\end{figure}

We study the impact of KD on student performance from two perspectives:
\textbf{(a)} \textit{Absolute reduction in student errors:}
Fig~\ref{fig:error_comparison} shows a class-wise comparison of the number of misclassified samples by the student with and without KD. Across both labels 0 \& 1, KD consistently reduces the total number of prediction errors, demonstrating its effectiveness in improving student generalization.  \textbf{(b)} \textit{Mitigation of errors where the teacher was correct:}
To assess whether the student learns from the teacher's correct predictions, we isolate those samples where the teacher was correct but the student was not. Fig~\ref{fig:mismatch_with_teacher} presents these student-teacher mismatches, which effectively represent cases where distillation has the potential to guide the student. After KD, the number of such mismatches drops substantially across both classes, showing that the student better aligns with the teacher’s correct decisions.

\begin{table}[t]
\centering

\small
\label{tab:cross_dataset_perf}
\renewcommand{\arraystretch}{1.3}
\setlength{\tabcolsep}{6pt}
\begin{adjustbox}{width=\linewidth}
\begin{tabular}{lcccc}
\toprule
\textbf{Method} & \multicolumn{2}{c}{\textbf{MCMD}} & \multicolumn{2}{c}{\textbf{RedEval}} \\
 & \textbf{Acc} & \textbf{F1} & \textbf{Acc} & \textbf{F1} \\
\midrule
\multicolumn{5}{c}{\cellcolor{gray!10}\textbf{Multimodal (Non-PEFT)}} \\
Att-BERT & 0.477 (0.3) & 0.474 (0.1) & 0.461 (1.1) & 0.457 (0.8) \\
DIP & 0.545 (1.2) & 0.545 (0.8) & 0.532 (0.2) & 0.513 (0.7) \\
DynRT & 0.519 (1.6) & 0.518 (1.4) & 0.541 (0.6) & 0.537 (0.9) \\
MV-CLIP & 0.653 (0.2) & 0.641 (0.2) & 0.623 (1.1) & 0.620 (1.3) \\
RAG-LLaVA & 0.624 (1.7) & 0.623 (1.5) & 0.617 (1.9) & 0.611 (1.4) \\
\midrule
\multicolumn{5}{c}{\cellcolor{gray!10}\textbf{Multimodal (PEFT)}} \\
PVLM & 0.564 (1.8) & 0.541 (1.3) & 0.553 (1.2) & 0.552 (1.1) \\
UP-MPF & 0.582 (2.1) & 0.577 (1.9) & 0.569 (0.1) & 0.561 (0.3) \\
CoOp  & 0.663 (0.4) & 0.662 (0.2) & 0.629 (0.9) & 0.618 (0.7) \\
CoCoOp  & 0.658 (1.0) & 0.649 (0.5) & 0.637 (1.2) & 0.631 (1.1) \\
CAMP & 0.601 (1.3) & 0.591 (1.6) & 0.631 (0.7) & 0.628 (1.2) \\
\midrule
\multicolumn{5}{c}{\cellcolor{gray!10}\textbf{Large Vision-Language Models (LVLMs)}} \\
LLaMA3.2-11B & 0.693 (0.4) & 0.682 (0.3) & 0.641 (0.1) & 0.653 (0.4) \\
LLaVA1.6-7B & 0.674 (0.5) & 0.669 (0.3) & 0.642 (0.4) & 0.641 (0.6) \\
Qwen2.5-VL-7B & 0.691 (0.3) & 0.685 (0.6) & 0.656 (0.7) & 0.659 (0.5) \\
\midrule
\multicolumn{5}{c}{\cellcolor{gray!10}\textbf{PEKD (Ours)}} \\
Prompt-CLIP (w/ KD) & 0.713 (0.1) & 0.698 (0.2) & 0.686 (0.4) & 0.683 (0.1) \\
Adapter-CLIP (w/ KD) & 0.724 (0.2) & 0.719 (0.3) & 0.698 (0.2) & 0.691 (0.4) \\
LoRA-CLIP (w/ KD) & \textbf{0.742 (0.1)} & \textbf{0.739 (0.2)} & \textbf{0.704 (0.2)} & \textbf{0.698 (0.4)} \\
\bottomrule
\end{tabular}
\end{adjustbox}
\caption{\label{tab:cross_dataset_eval}
Cross-dataset evaluation. All models are trained on 1\% of MMSD2.0 and tested on two unseen datasets: MCMD and RedEval.  PEKD consistently boosts all PEFT variants. Numbers in brackets denote standard deviation. Best results are in \textbf{bold}.}
\end{table}

\begingroup
\tiny
\begin{table}[t]
\centering
\renewcommand{\arraystretch}{1.0}
\setlength{\tabcolsep}{2pt}
\begin{adjustbox}{width=0.99\columnwidth}
\begin{tabular}{cc}
\begin{minipage}[t]{0.48\columnwidth}
\centering
\includegraphics[width=2.5cm]{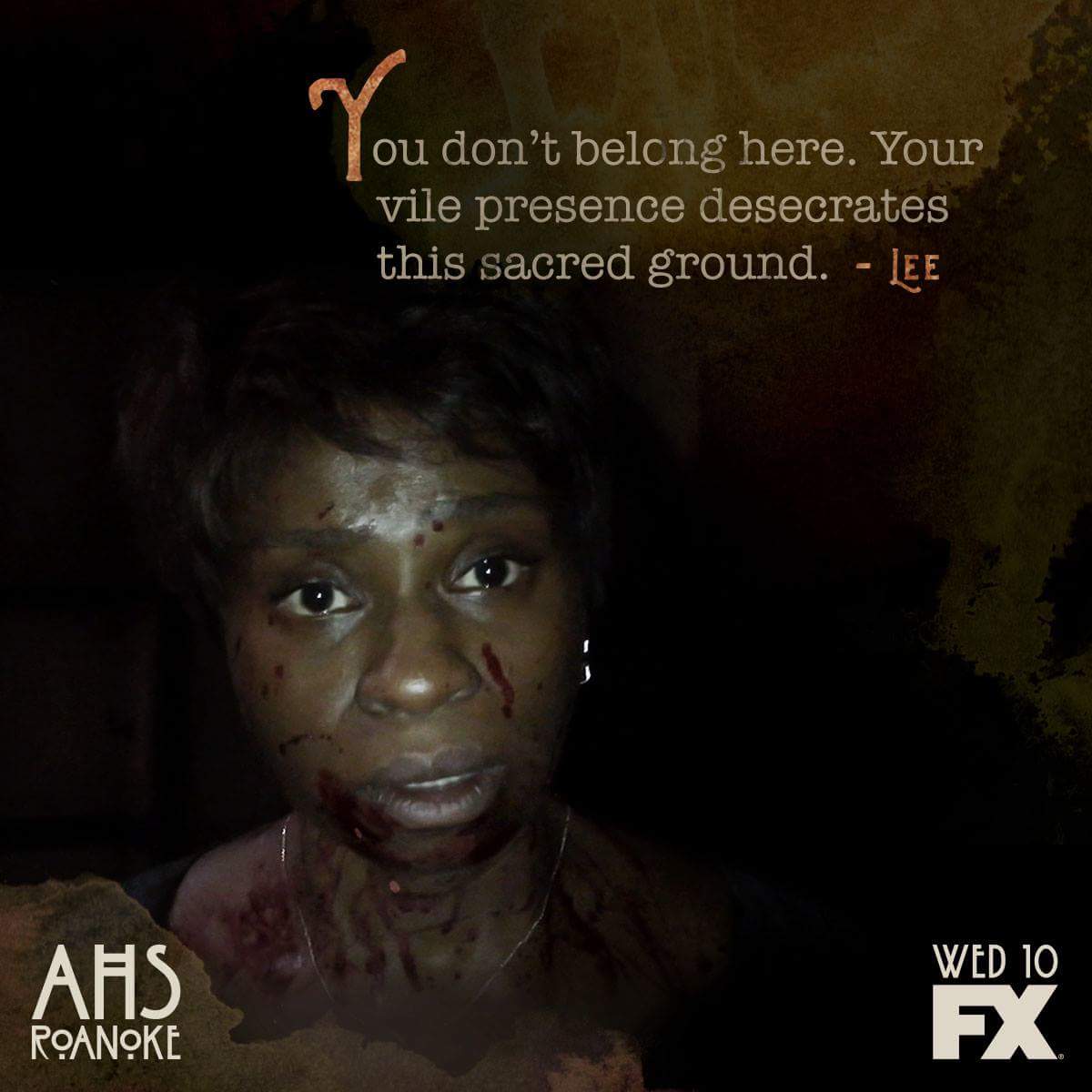} \\
\vspace{2pt}
\footnotesize (a) what i plan on saying to anyone that rings my doorbell \\
\vspace{1pt}
\scriptsize \textbf{GT / T / S:} 1 / 0 / 0
\end{minipage}
&
\begin{minipage}[t]{0.48\columnwidth}
\centering
\includegraphics[width=2.5cm]{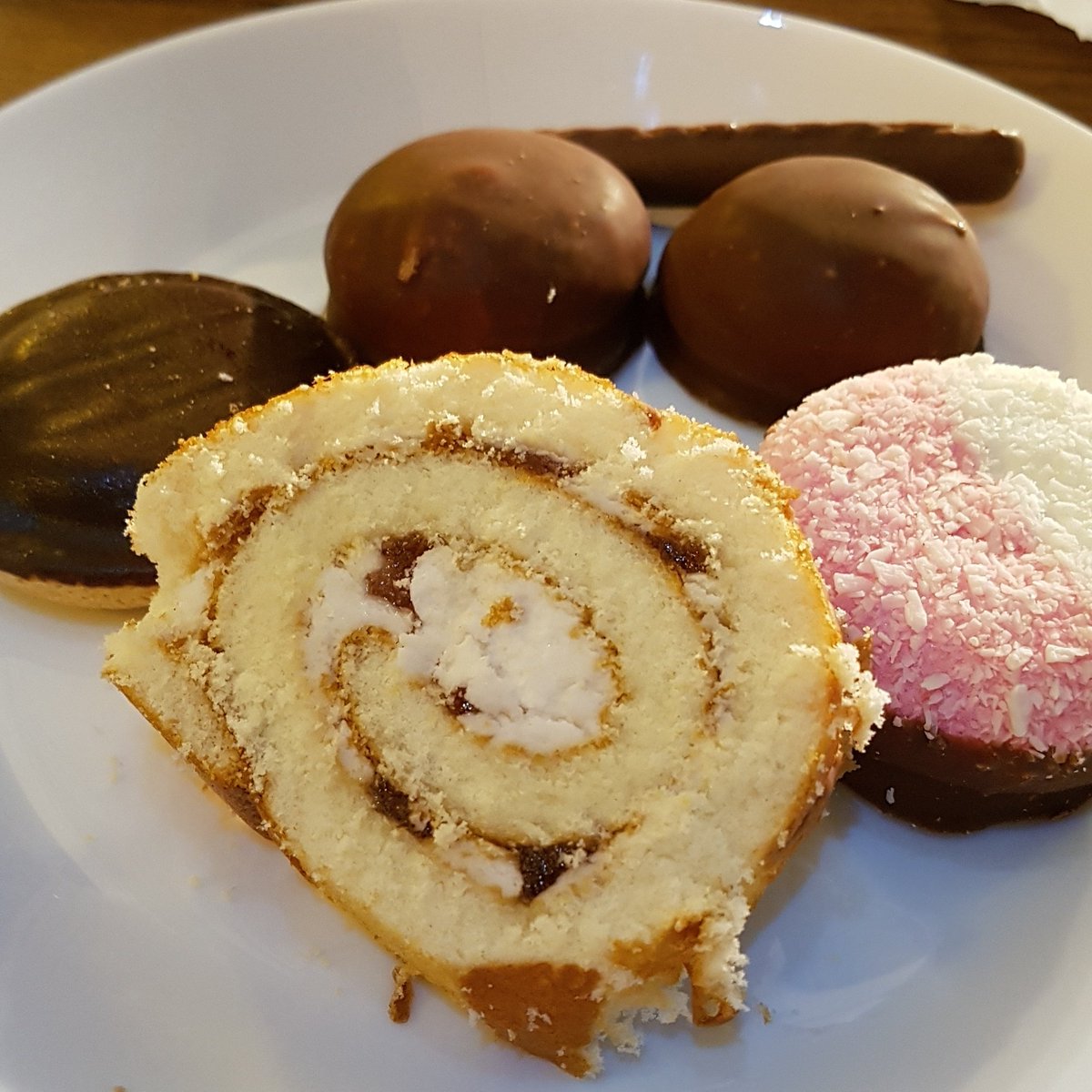} \\
\vspace{2pt}
\footnotesize (b) authentic chinese desserts \\
\vspace{1pt}
\scriptsize \textbf{GT / T / S:} 1 / 0 / 0
\end{minipage}
\\[8pt]

\begin{minipage}[t]{0.48\columnwidth}
\centering
\includegraphics[width=2.5cm]{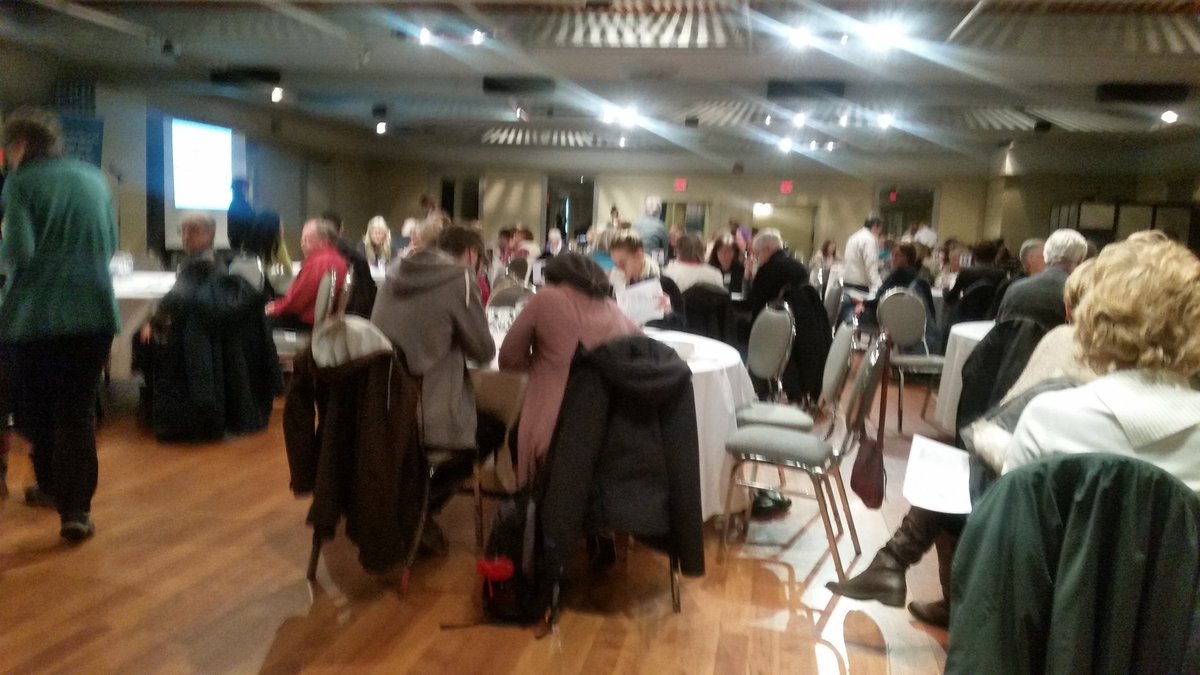} \\
\vspace{2pt}
\footnotesize (c) capacity room at basic income consultation in kingston \\
\vspace{1pt}
\scriptsize \textbf{GT / T / S:} 0 / 0 / 1
\end{minipage}
&
\begin{minipage}[t]{0.48\columnwidth}
\centering
\includegraphics[width=2.5cm]{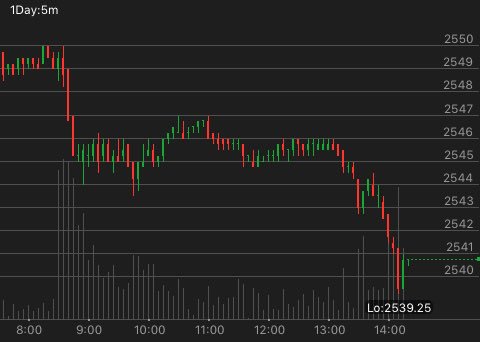} \\
\vspace{8pt}
\footnotesize (d) major pullback ! ! 5pts \\
\vspace{1pt}
\scriptsize \textbf{GT / T / S:} 1 / 1 / 0
\end{minipage}
\end{tabular}
\end{adjustbox}
\caption{\label{tab:errors}
Qualitative error examples with image, caption, and prediction (GT: ground truth, T: teacher, S: student). Refer to appendix \ref{app:error_details} for an explanation of the errors.}
\end{table}
\endgroup

\subsection{Cross-Dataset Generalization}
To evaluate the generalizability of our proposed framework, we conduct cross-dataset experiments by training all models on just 1\% of the MMSD2.0 dataset and testing them on two unseen datasets: MCMD \cite{Maity2022AMF} and RedEval \cite{Tang2024LeveragingGL}. The details of these datasets are provided in Appendix \ref{app:cross_dataset}. As shown in Table~\ref{tab:cross_dataset_eval}, PEKD-based models consistently outperform both PEFT and non-PEFT baselines, and even LVLMs across both datasets. Notably, the best-performing PEKD variant (\texttt{LoRA-CLIP}) achieves a substantial improvement over the strongest baseline (\texttt{Qwen2.5-VL-7B}). These results prove the generalizability of PEKD framework.

\section{Error Analysis}
To better understand the failure cases of our PEKD framework, we qualitatively analyze prediction errors and identify two broad patterns, shown in Table \ref{tab:errors}. First, both teacher and student models fail on (i) \textbf{OCR-heavy cases}, where sarcasm cues are embedded as image text (e.g., memes, screenshots), and (ii) \textbf{context-dependent cases}, where interpretation relies on external knowledge such as political or cultural context, inaccessible to the models (row 1). Second, we observe \textbf{student-specific failures} where the teacher predicts correctly but the student does not—often due to limited data exposure or weaker multimodal reasoning capacity (row 2).

\section{Conclusion}
In this work, we introduce \textbf{PEKD}, a framework for few-shot multimodal sarcasm detection that combines parameter-efficient tuning with knowledge distillation from a strong CLIP-based teacher. By guiding lightweight CLIP student variants through KD and incorporating an entropy-aware gating mechanism to prioritize reliable teacher signals, PEKD enhances robustness and performance in few-shot settings. Experiments on two benchmarks show consistent gains over parameter-efficient and large multimodal baselines, highlighting its potential for scalable vision-language understanding under data scarcity.

\section{Limitations}
While PEKD delivers notable improvements in few-shot multimodal sarcasm detection, it has some limitations. Both teacher and student primarily leverage visual-textual content and can struggle in OCR-heavy scenarios or cases requiring external context. Lastly, the entropy-based weighting of the KD signal is a simple confidence proxy and may not fully capture nuanced uncertainties of sarcasm. Future work could explore richer reliability estimation and integration of external knowledge to improve robustness and reasoning.

\bibliography{custom}

\begin{thebibliography}{35}
\expandafter\ifx\csname natexlab\endcsname\relax\def\natexlab#1{#1}\fi

\bibitem[{Bai et~al.(2025)Bai, Chen, Liu, Wang, Ge, Song, Dang, Wang, Wang, Tang, Zhong, Zhu, Yang, Li, Wan, Wang, Ding, Fu, Xu, Ye, Zhang, Xie, Cheng, Zhang, Yang, Xu, and Lin}]{Bai2025Qwen25VLTR}
Shuai Bai, Keqin Chen, Xuejing Liu, Jialin Wang, Wenbin Ge, Sibo Song, Kai Dang, Peng Wang, Shijie Wang, Jun Tang, Humen Zhong, Yuanzhi Zhu, Mingkun Yang, Zhaohai Li, Jianqiang Wan, Pengfei Wang, Wei Ding, Zheren Fu, Yiheng Xu, Jiabo Ye, Xi~Zhang, Tianbao Xie, Zesen Cheng, Hang Zhang, Zhibo Yang, Haiyang Xu, and Junyang Lin. 2025.
\newblock Qwen2.5-vl technical report.
\newblock \emph{ArXiv}.

\bibitem[{Cai et~al.(2019)Cai, Cai, and Wan}]{cai-etal-2019-multi}
Yitao Cai, Huiyu Cai, and Xiaojun Wan. 2019.
\newblock Multi-modal sarcasm detection in {T}witter with hierarchical fusion model.
\newblock In \emph{Proceedings of the 57th Annual Meeting of the Association for Computational Linguistics}, Florence, Italy. Association for Computational Linguistics.

\bibitem[{Davidov et~al.(2010)Davidov, Tsur, and Rappoport}]{davidov-etal-2010-semi}
Dmitry Davidov, Oren Tsur, and Ari Rappoport. 2010.
\newblock Semi-supervised recognition of sarcasm in {T}witter and {A}mazon.
\newblock In \emph{Proceedings of the Fourteenth Conference on Computational Natural Language Learning}, Uppsala, Sweden. Association for Computational Linguistics.

\bibitem[{Dress et~al.(2008)Dress, Kreuz, Link, and Caucci}]{Dress2008RegionalVI}
Megan~L. Dress, Roger~J. Kreuz, Kristen~E. Link, and Gina~M. Caucci. 2008.
\newblock Regional variation in the use of sarcasm.
\newblock \emph{Journal of Language and Social Psychology}, 27:71 -- 85.

\bibitem[{et~al.(2024)}]{Dubey2024TheL3}
Abhimanyu~Dubey et~al. 2024.
\newblock The llama 3 herd of models.
\newblock \emph{ArXiv}.

\bibitem[{Gao et~al.(2021)Gao, Geng, Zhang, Ma, Fang, Zhang, Li, and Qiao}]{Gao2021CLIPAdapterBV}
Peng Gao, Shijie Geng, Renrui Zhang, Teli Ma, Rongyao Fang, Yongfeng Zhang, Hongsheng Li, and Yu~Jiao Qiao. 2021.
\newblock Clip-adapter: Better vision-language models with feature adapters.
\newblock \emph{ArXiv}.

\bibitem[{Gonz{\'a}lez-Ib{\'a}{\~n}ez et~al.(2011)Gonz{\'a}lez-Ib{\'a}{\~n}ez, Muresan, and Wacholder}]{GonzlezIbez2011IdentifyingSI}
Roberto~I. Gonz{\'a}lez-Ib{\'a}{\~n}ez, Smaranda Muresan, and Nina Wacholder. 2011.
\newblock Identifying sarcasm in twitter: A closer look.
\newblock In \emph{Annual Meeting of the Association for Computational Linguistics}.

\bibitem[{Hinton et~al.(2015)Hinton, Vinyals, and Dean}]{Hinton2015DistillingTK}
Geoffrey~E. Hinton, Oriol Vinyals, and Jeffrey Dean. 2015.
\newblock Distilling the knowledge in a neural network.
\newblock \emph{ArXiv}.

\bibitem[{Houlsby et~al.(2019)Houlsby, Giurgiu, Jastrzebski, Morrone, de~Laroussilhe, Gesmundo, Attariyan, and Gelly}]{Houlsby2019ParameterEfficientTL}
Neil Houlsby, Andrei Giurgiu, Stanislaw Jastrzebski, Bruna Morrone, Quentin de~Laroussilhe, Andrea Gesmundo, Mona Attariyan, and Sylvain Gelly. 2019.
\newblock Parameter-efficient transfer learning for nlp.
\newblock \emph{ArXiv}.

\bibitem[{Hu et~al.(2021)Hu, Shen, Wallis, Allen-Zhu, Li, Wang, and Chen}]{Hu2021LoRALA}
J.~Edward Hu, Yelong Shen, Phillip Wallis, Zeyuan Allen-Zhu, Yuanzhi Li, Shean Wang, and Weizhu Chen. 2021.
\newblock Lora: Low-rank adaptation of large language models.
\newblock \emph{ArXiv}.

\bibitem[{Jana et~al.(2024)Jana, Dey, and Sanasam}]{Jana2024ContinuousAM}
Soumyadeep Jana, Animesh Dey, and Ranbir~Singh Sanasam. 2024.
\newblock Continuous attentive multimodal prompt tuning for few-shot multimodal sarcasm detection.
\newblock \emph{Proceedings of the 28th Conference on Computational Natural Language Learning}.

\bibitem[{Lester et~al.(2021)Lester, Al-Rfou, and Constant}]{Lester2021ThePO}
Brian Lester, Rami Al-Rfou, and Noah Constant. 2021.
\newblock The power of scale for parameter-efficient prompt tuning.
\newblock In \emph{Conference on Empirical Methods in Natural Language Processing}.

\bibitem[{Liang et~al.(2021)Liang, Lou, Li, Gui, Yang, and Xu}]{Liang2021MultiModalSD}
Bin Liang, Chenwei Lou, Xiang Li, Lin Gui, Min Yang, and Ruifeng Xu. 2021.
\newblock Multi-modal sarcasm detection with interactive in-modal and cross-modal graphs.
\newblock \emph{Proceedings of the 29th ACM International Conference on Multimedia}.

\bibitem[{Liang et~al.(2022{\natexlab{a}})Liang, Lou, Li, Yang, Gui, He, Pei, and Xu}]{Liang2022MultiModalSD}
Bin Liang, Chenwei Lou, Xiang Li, Min Yang, Lin Gui, Yulan He, Wenjie Pei, and Ruifeng Xu. 2022{\natexlab{a}}.
\newblock Multi-modal sarcasm detection via cross-modal graph convolutional network.
\newblock In \emph{Annual Meeting of the Association for Computational Linguistics}.

\bibitem[{Liang et~al.(2022{\natexlab{b}})Liang, Zhao, and Sch{\"u}tze}]{Liang2022ModularAP}
Sheng Liang, Mengjie Zhao, and Hinrich Sch{\"u}tze. 2022{\natexlab{b}}.
\newblock Modular and parameter-efficient multimodal fusion with prompting.
\newblock \emph{ArXiv}.

\bibitem[{Liu et~al.(2023)Liu, Li, Li, and Lee}]{Liu2023ImprovedBW}
Haotian Liu, Chunyuan Li, Yuheng Li, and Yong~Jae Lee. 2023.
\newblock Improved baselines with visual instruction tuning.
\newblock \emph{2024 IEEE/CVF Conference on Computer Vision and Pattern Recognition (CVPR)}.

\bibitem[{Liu et~al.(2022)Liu, Wang, and Li}]{liu-etal-2022-towards-multi-modal}
Hui Liu, Wenya Wang, and Haoliang Li. 2022.
\newblock Towards multi-modal sarcasm detection via hierarchical congruity modeling with knowledge enhancement.
\newblock In \emph{Proceedings of the 2022 Conference on Empirical Methods in Natural Language Processing}, Abu Dhabi, United Arab Emirates. Association for Computational Linguistics.

\bibitem[{Maity et~al.(2022)Maity, Jha, Saha, and Bhattacharyya}]{Maity2022AMF}
Krishanu Maity, Prince Jha, Sriparna Saha, and Pushpak Bhattacharyya. 2022.
\newblock A multitask framework for sentiment, emotion and sarcasm aware cyberbullying detection from multi-modal code-mixed memes.
\newblock \emph{Proceedings of the 45th International ACM SIGIR Conference on Research and Development in Information Retrieval}.

\bibitem[{Oprea and Magdy(2019)}]{oprea-magdy-2019-exploring}
Silviu Oprea and Walid Magdy. 2019.
\newblock Exploring author context for detecting intended vs perceived sarcasm.
\newblock In \emph{Proceedings of the 57th Annual Meeting of the Association for Computational Linguistics}, Florence, Italy. Association for Computational Linguistics.

\bibitem[{Pan et~al.(2020)Pan, Lin, Fu, Qi, and Wang}]{pan-etal-2020-modeling}
Hongliang Pan, Zheng Lin, Peng Fu, Yatao Qi, and Weiping Wang. 2020.
\newblock Modeling intra and inter-modality incongruity for multi-modal sarcasm detection.
\newblock In \emph{Findings of the Association for Computational Linguistics: EMNLP 2020}, Online. Association for Computational Linguistics.

\bibitem[{Qin et~al.(2023)Qin, Huang, Chen, Cai, Zhang, Liang, Che, and Xu}]{qin-etal-2023-mmsd2}
Libo Qin, Shijue Huang, Qiguang Chen, Chenran Cai, Yudi Zhang, Bin Liang, Wanxiang Che, and Ruifeng Xu. 2023.
\newblock {MMSD}2.0: Towards a reliable multi-modal sarcasm detection system.
\newblock In \emph{Findings of the Association for Computational Linguistics: ACL 2023}, Toronto, Canada. Association for Computational Linguistics.

\bibitem[{Radford et~al.(2021)Radford, Kim, Hallacy, Ramesh, Goh, Agarwal, Sastry, Askell, Mishkin, Clark, Krueger, and Sutskever}]{Radford2021LearningTV}
Alec Radford, Jong~Wook Kim, Chris Hallacy, Aditya Ramesh, Gabriel Goh, Sandhini Agarwal, Girish Sastry, Amanda Askell, Pamela Mishkin, Jack Clark, Gretchen Krueger, and Ilya Sutskever. 2021.
\newblock Learning transferable visual models from natural language supervision.
\newblock In \emph{International Conference on Machine Learning}.

\bibitem[{Rockwell and Theriot(2001)}]{Rockwell2001CultureGA}
Patricia Rockwell and Evelyn~M. Theriot. 2001.
\newblock Culture, gender, and gender mix in encoders of sarcasm: A self‐assessment analysis.
\newblock \emph{Communication Research Reports}, 18:44 -- 52.

\bibitem[{Schifanella et~al.(2016)Schifanella, de~Juan, Tetreault, and Cao}]{Schifanella2016DetectingSI}
Rossano Schifanella, Paloma de~Juan, Joel~R. Tetreault, and Liangliang Cao. 2016.
\newblock Detecting sarcasm in multimodal social platforms.
\newblock \emph{Proceedings of the 24th ACM international conference on Multimedia}.

\bibitem[{Tang et~al.(2024)Tang, Lin, Yan, and Li}]{Tang2024LeveragingGL}
Binghao Tang, Boda Lin, Haolong Yan, and Si~Li. 2024.
\newblock Leveraging generative large language models with visual instruction and demonstration retrieval for multimodal sarcasm detection.
\newblock In \emph{North American Chapter of the Association for Computational Linguistics}.

\bibitem[{Tian et~al.(2023)Tian, Xu, Zhang, and Mao}]{tian-etal-2023-dynamic}
Yuan Tian, Nan Xu, Ruike Zhang, and Wenji Mao. 2023.
\newblock Dynamic routing transformer network for multimodal sarcasm detection.
\newblock In \emph{Proceedings of the 61st Annual Meeting of the Association for Computational Linguistics (Volume 1: Long Papers)}, Toronto, Canada. Association for Computational Linguistics.

\bibitem[{Wen et~al.(2023)Wen, Jia, and Yang}]{Wen2023DIPDI}
Chan~Shao Wen, Guoli Jia, and Jufeng Yang. 2023.
\newblock Dip: Dual incongruity perceiving network for sarcasm detection.
\newblock \emph{2023 IEEE/CVF Conference on Computer Vision and Pattern Recognition (CVPR)}.

\bibitem[{Wu et~al.(2024)Wu, Huang, Qu, and Wu}]{Wu2024MixtureofPromptExpertsFM}
Zichen Wu, Hsiu-Yuan Huang, Fanyi Qu, and Yunfang Wu. 2024.
\newblock Mixture-of-prompt-experts for multi-modal semantic understanding.
\newblock In \emph{International Conference on Language Resources and Evaluation}.

\bibitem[{Xie et~al.(2024)Xie, Zhu, Chen, Chen, and Huang}]{10.1145/3664647.3680914}
Yifeng Xie, Zhihong Zhu, Xin Chen, Zhanpeng Chen, and Zhiqi Huang. 2024.
\newblock Moba: Mixture of bi-directional adapter for multi-modal sarcasm detection.
\newblock In \emph{Proceedings of the 32nd ACM International Conference on Multimedia}. Association for Computing Machinery.

\bibitem[{Xu et~al.(2020)Xu, Zeng, and Mao}]{xu-etal-2020-reasoning}
Nan Xu, Zhixiong Zeng, and Wenji Mao. 2020.
\newblock Reasoning with multimodal sarcastic tweets via modeling cross-modality contrast and semantic association.
\newblock In \emph{Proceedings of the 58th Annual Meeting of the Association for Computational Linguistics}, Online. Association for Computational Linguistics.

\bibitem[{Yang et~al.(2022)Yang, Feng, Wang, Hong, and Poria}]{Yang2022FewshotMS}
Xiaocui Yang, Shi Feng, Daling Wang, Pengfei Hong, and Soujanya Poria. 2022.
\newblock Few-shot multimodal sentiment analysis based on multimodal probabilistic fusion prompts.
\newblock \emph{Proceedings of the 31st ACM International Conference on Multimedia}.

\bibitem[{Yu and Zhang(2022)}]{Yu2022FewShotMS}
Yang Yu and Dong Zhang. 2022.
\newblock Few-shot multi-modal sentiment analysis with prompt-based vision-aware language modeling.
\newblock \emph{2022 IEEE International Conference on Multimedia and Expo (ICME)}.

\bibitem[{Yu et~al.(2022)Yu, Zhang, and Li}]{Yu2022UnifiedMP}
Yang Yu, Dong Zhang, and Shoushan Li. 2022.
\newblock Unified multi-modal pre-training for few-shot sentiment analysis with prompt-based learning.
\newblock \emph{Proceedings of the 30th ACM International Conference on Multimedia}.

\bibitem[{Zhou et~al.(2021)Zhou, Yang, Loy, and Liu}]{Zhou2021LearningTP}
Kaiyang Zhou, Jingkang Yang, Chen~Change Loy, and Ziwei Liu. 2021.
\newblock Learning to prompt for vision-language models.
\newblock \emph{International Journal of Computer Vision}.

\bibitem[{Zhou et~al.(2022)Zhou, Yang, Loy, and Liu}]{Zhou2022ConditionalPL}
Kaiyang Zhou, Jingkang Yang, Chen~Change Loy, and Ziwei Liu. 2022.
\newblock Conditional prompt learning for vision-language models.
\newblock \emph{2022 IEEE/CVF Conference on Computer Vision and Pattern Recognition (CVPR)}.

\end{thebibliography}
\bibliographystyle{acl_natbib}

\appendix

\section{Appendix}
\label{sec:appendix}

\subsection{Hyperparameter Details}
\label{app:hyperparameter}
The teacher was fine‑tuned till the best val accuracy was reached, with learning rates of 1e‑5 for the backbone and 1e‑4 for the head and a batch size of 32. The student was fine-tuned with a learning rate of 1e‑4, batch size of 32, with a LoRA rank of 32. The best model on the validation set was used for testing. The value of temperature $T$ for the KD loss was set to 2. All experiments were conducted on an Nvidia RTX A5000 GPU with 24GB of memory. 

\subsection{Baseline Details}
\label{app:baselines}
To establish a comprehensive comparison, we evaluate our method against two categories of multimodal sarcasm detection baselines: (i) \textbf{Non-PEFT-based methods}, which primarily focus on modeling image-text interactions through hierarchical fusion, graph networks, co-attention, or dynamic routing, and (ii) \textbf{PEFT-based methods}, which leverage prompts and adapters to address few-shot multimodal sentiment, sarcasm and image recognition tasks. 

\textbf{Non-PEFT Baselines:}  
\begin{itemize}
    \item \textbf{HFM} \cite{cai-etal-2019-multi}: Introduces hierarchical early and late fusion to integrate image, text, and image attributes.
    \item \textbf{D\&R Net} \cite{xu-etal-2020-reasoning}: Uses decomposition and relational reasoning to capture semantic associations.
    \item \textbf{Att-BERT} \cite{pan-etal-2020-modeling}: Employs co-attention to detect intra- and inter-modal incongruity.
    \item \textbf{HKE} \cite{liu-etal-2022-towards-multi-modal}: Leverages a hierarchical network to model coarse- and fine-grained incongruities.
    \item \textbf{MV-CLIP} \cite{qin-etal-2023-mmsd2}: Extends CLIP with an interaction layer for enhanced image-text incongruity modeling.
    \item \textbf{DIP} \cite{Wen2023DIPDI}: Captures sarcasm via semantic reweighting, uncertainty modeling, and contrastive learning at factual and affective levels.
    \item \textbf{DynRT} \cite{tian-etal-2023-dynamic}: Employs dynamic routing to identify sarcasm-relevant tokens in multimodal data.
    \item \textbf{RAG-LLaVA} \cite{Tang2024LeveragingGL}: Incorporates retrieval-based demonstrations to assist LLaVA for sarcasm detection.
\end{itemize}

\textbf{PEFT baselines:}  
\begin{itemize}
    \item \textbf{PVLM} \cite{Yu2022FewShotMS}: Adopts prompt-based fine-tuning with integrated image tokens for few-shot multimodal sentiment analysis.
    \item \textbf{UP-MPF} \cite{Yu2022UnifiedMP}: Pretrains on image tasks to bridge the gap between textual and visual prompts before applying to few-shot sentiment tasks.
    \item \textbf{CoOp} \cite{Zhou2021LearningTP}: Learns continuous text prompts for CLIP for image recognition.
    \item \textbf{CoCoOp} \cite{Zhou2022ConditionalPL}: Introduces image-conditioned text prompts for improved adaptation for CLIP for image recognition.
    \item \textbf{CAMP} \cite{Jana2024ContinuousAM}: Proposes continuous attentive prompt tokens in BERT for few-shot multimodal sarcasm detection.
    \item \textbf{MoBA} \cite{10.1145/3664647.3680914}: Proposed mixture of adapters-based technique to combine visual and textual information for multimodal sarcasm detection, achieving parameter efficiency.

\end{itemize}

\subsection{Cross-Dataset Details}
\label{app:cross_dataset}
To evaluate the generalization ability of multimodal sarcasm detection models, we introduce two out-of-distribution datasets: \textbf{MCMD} and \textbf{RedEval}. 

\begin{itemize}
    \item \textbf{MCMD (Multi-modal Code-Mixed Memes Dataset)} \cite{Maity2022AMF}: This dataset consists of code-mixed memes originally meant for hateful meme detection but also has annotations for sarcasm. We remove samples that are code-mixed in nature or has no sarcasm annotations.
    
    \item \textbf{RedEval} \cite{Tang2024LeveragingGL}: Constructed to assess domain shift, RedEval includes image-text pairs from \textit{Reddit}, since existing MMSD datasets are Twitter-centric. Sarcastic examples are sourced from the \textit{sarcasm} subreddit, while non-sarcastic examples come from subreddits like \textit{aww}, \textit{funny}, and \textit{pics}.

    The detailed statistics for both these datasets are in Table \ref{tab:cross_data}
\end{itemize}

\begingroup
\renewcommand{\arraystretch}{1.2}
\begin{table}[t]
\centering
\scriptsize
\setlength{\tabcolsep}{6pt}
\begin{tabular}{lccc}
\toprule
\textbf{Dataset} & \textbf{Sarcastic} & \textbf{Non-Sarcastic} & \textbf{Total} \\
\midrule
\textbf{MCMD}  & 183 & 123 & 306 \\
\textbf{RedEval} & 395 & 609 & 1004 \\
\bottomrule
\end{tabular}
\caption{\label{tab:cross_data}
Statistics of datasets used for cross-dataset evaluation.}
\end{table}
\endgroup

\subsection{Proof for Lemma 1}
\label{app:proof}
\begin{proof}
The entropy-aware gating factor is defined as:
\begin{align}
g = \frac{H(y_\mathcal{T})}{\log C}, \quad 
H(y_\mathcal{T}) = -\sum_{i=1}^{C} y_\mathcal{T}(i)\log y_\mathcal{T}(i),
\end{align}
where $\log C$ is the maximum possible entropy for $C$ classes, normalizing $g \in [0,1]$.  
The gated KD loss:
\begin{align}
\mathcal{L}_{KD}^{gated} = (1-g)\,\mathcal{L}_{KD}.
\end{align}

\textbf{Case 1: Teacher is highly uncertain.}  
If $y_\mathcal{T}(i) \approx \frac{1}{C}$ for all $i$, then:
\begin{align}
H(y_\mathcal{T}) &= -\sum_{i=1}^{C}\tfrac{1}{C}\log\tfrac{1}{C} = \log C,\\
g &= \tfrac{\log C}{\log C}=1,\;\;\; \mathcal{L}_{KD}^{gated}=0.
\end{align}

\textbf{Case 2: Teacher is highly confident.}  
If teacher predicts one class with high probability:
\begin{align}
H(y_\mathcal{T}) \approx 0,\;\; g=0,\;\; \mathcal{L}_{KD}^{gated}=\mathcal{L}_{KD}.
\end{align}

\textbf{Conclusion:}  
Weighting by $(1-g)$ makes KD vanish when $g\to 1$ (uncertain teacher) and fully retain it when $g\to 0$ (confident teacher). This prevents noisy guidance while exploiting strong teacher signals.

\end{proof}

\begingroup
\renewcommand{\arraystretch}{1.2}
\begin{table}[t]
\centering
\scriptsize
\setlength{\tabcolsep}{3pt}
\begin{tabular}{lcccccc}
\toprule
\multirow{2}{*}{\textbf{Model}} & \multicolumn{3}{c}{\textbf{MMSD2.0}} & \multicolumn{3}{c}{\textbf{MMSD}} \\
\cmidrule(lr){2-4} \cmidrule(lr){5-7}
 & Hard-Gate & Entropy & $\Delta$ & Hard-Gate & Entropy & $\Delta$ \\
\midrule
Prompt-CLIP  & 0.775 & \textbf{0.781} & \textcolor{green!70!black}{+0.6\%} & 0.815 & \textbf{0.821} & \textcolor{green!70!black}{+0.6\%} \\
Adapter-CLIP & 0.791 & \textbf{0.799} & \textcolor{green!70!black}{+0.8\%} & 0.817 & \textbf{0.824} & \textcolor{green!70!black}{+0.7\%} \\
LoRA-CLIP    & 0.803 & \textbf{0.811} & \textcolor{green!70!black}{+0.8\%} & 0.834 & \textbf{0.845} & \textcolor{green!70!black}{+1.1\%} \\
\bottomrule
\end{tabular}
\caption{Comparison of KD Variants. Entropy-gated KD outperforms Hard-Gate KD across models and datasets. $\Delta$ shows accuracy gain.}
\label{tab:kd_variant_ablation}
\end{table}
\endgroup

\subsection{Effect of Hard-Gate on KD Loss}
\label{app:hard_gate}  
We experimented with an alternative KD strategy, \textbf{Hard-Gate KD}, which discards the KD loss whenever the teacher prediction is incorrect. Table~\ref{tab:kd_variant_ablation} compares this approach against our \textbf{Entropy-Gated KD}. The entropy-based approach consistently outperforms the hard-zero variant by \textbf{0.6--1.1\% in accuracy} across both datasets and all PEFT configurations.  
\textit{Reason:} Hard-Gate KD eliminates valuable soft-label information even in cases where the teacher prediction is slightly off but provides useful class-probability distribution. Entropy gating, on the other hand, dynamically reduces KD influence without discarding it entirely, leading to better knowledge transfer.

\subsection{Detailed Error Analysis}
\label{app:error_details}

\begin{figure}[h]
\centering
\renewcommand{\arraystretch}{1.0}
\setlength{\tabcolsep}{2pt}
\begin{adjustbox}{width=\columnwidth}
\begin{tabular}{cc}
\begin{minipage}[t]{0.48\columnwidth}
\centering
\includegraphics[width=0.9\linewidth]{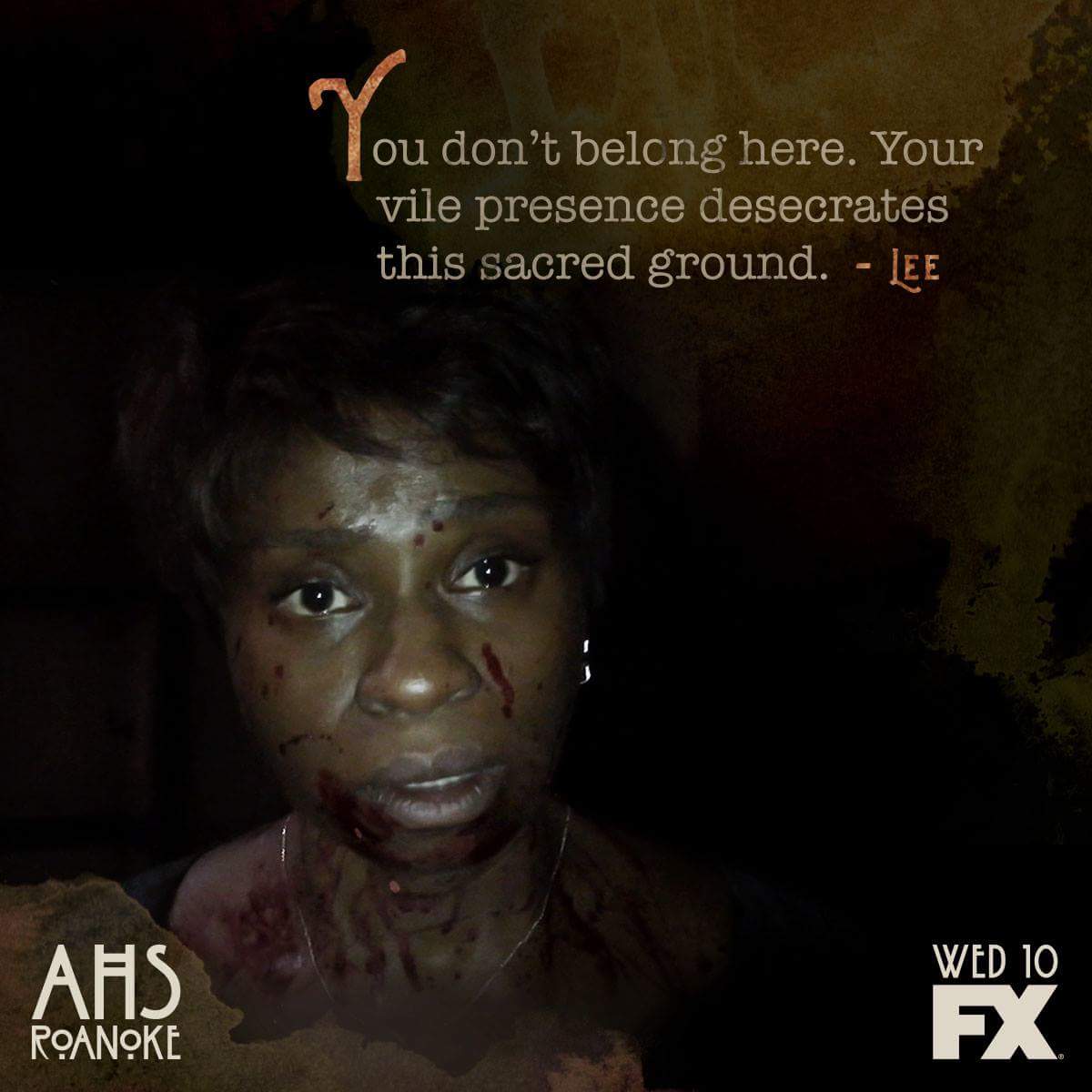} \\
\scriptsize \textbf{(a) OCR-heavy} \\
\scriptsize \emph{Caption:} ``what i plan on saying to anyone that rings my doorbell'' \\
\scriptsize \textbf{GT/T/S:} 1 / 0 / 0 \\
\scriptsize \textbf{Reason:} Sarcasm cue appears in text embedded in image; both models fail.
\end{minipage}
&
\begin{minipage}[t]{0.48\columnwidth}
\centering
\includegraphics[width=0.9\linewidth]{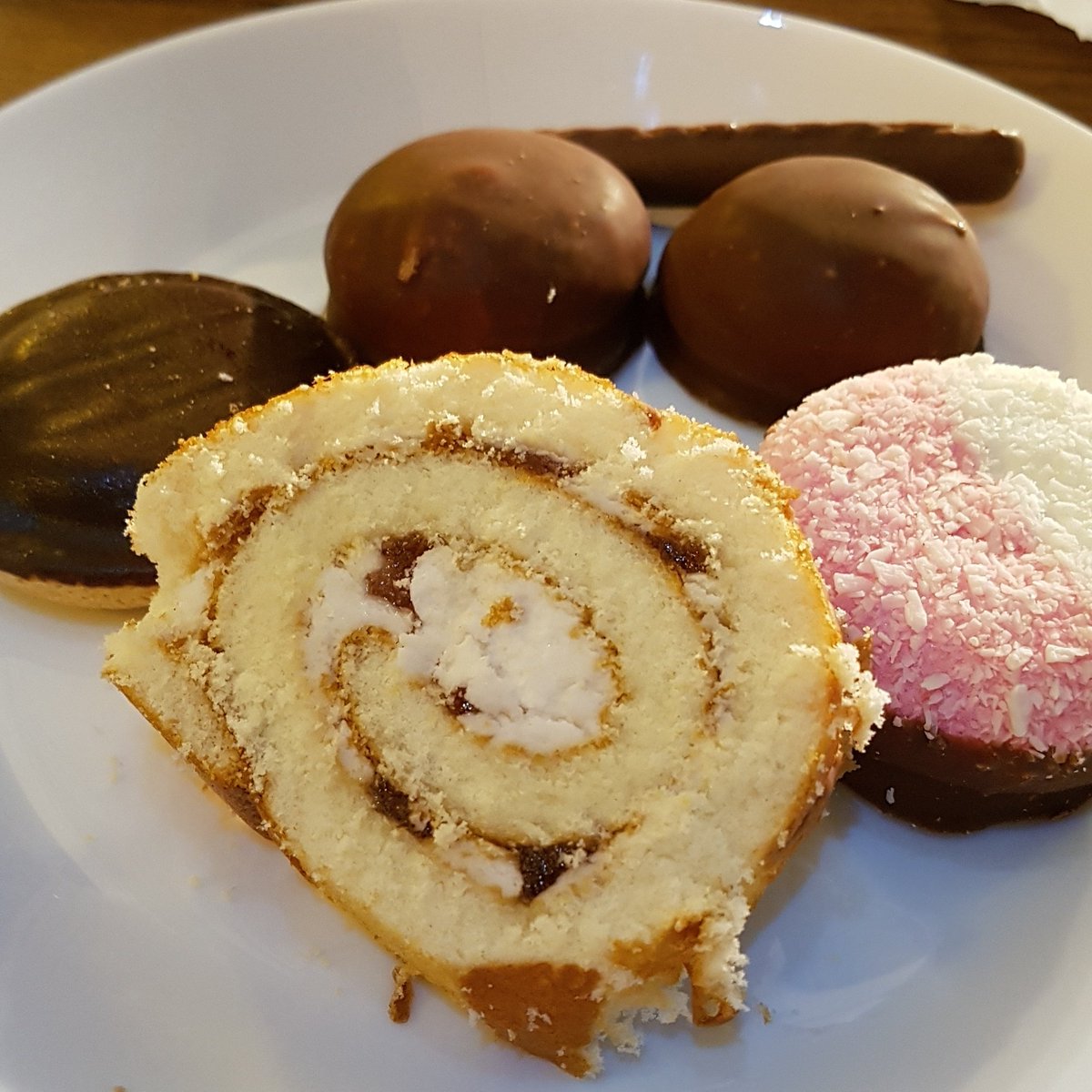} \\
\scriptsize \textbf{(b) Context-dependent} \\
\scriptsize \emph{Caption:} ``authentic chinese desserts'' \\
\scriptsize \textbf{GT/T/S:} 1 / 0 / 0 \\
\scriptsize \textbf{Reason:} Requires cultural knowledge as this dessert is not Chinese; both models fail.
\end{minipage}
\\[8pt]

\begin{minipage}[t]{0.48\columnwidth}
\centering
\includegraphics[width=0.9\linewidth]{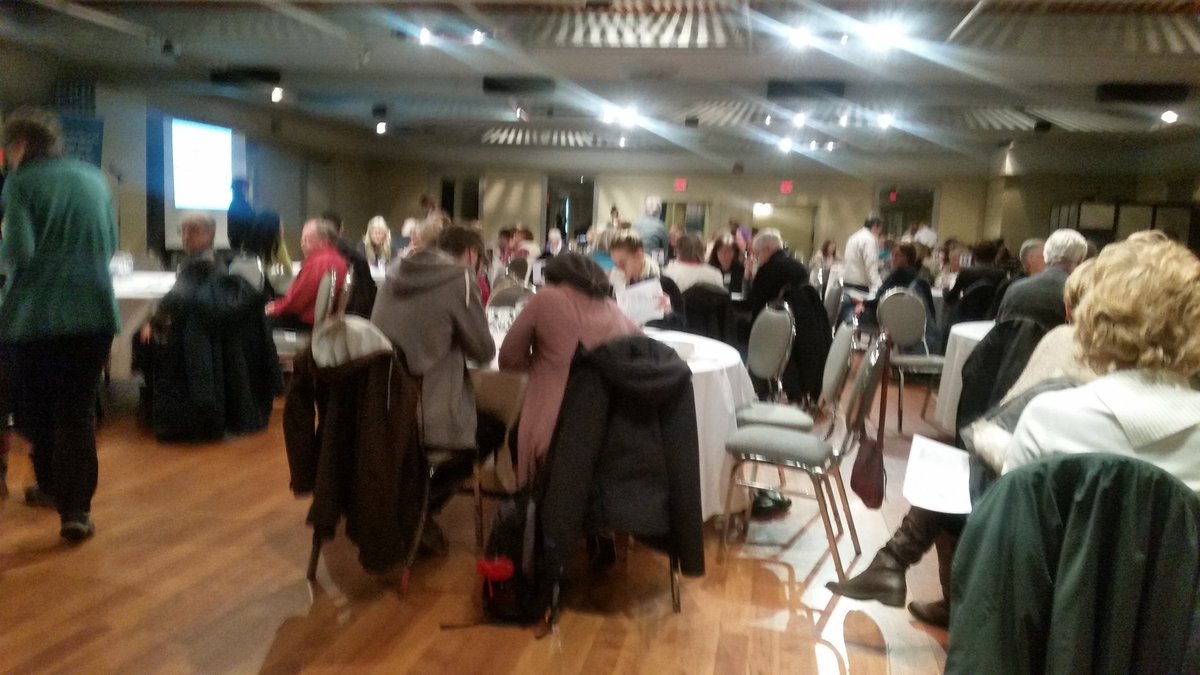} \\
\scriptsize \textbf{(c) Student-only} \\
\scriptsize \emph{Caption:} ``capacity room at basic income consultation'' \\
\scriptsize \textbf{GT/T/S:} 0 / 0 / 1 \\
\scriptsize \textbf{Reason:} Student overfits sarcastic patterns, might think that sparse room contradicts caption.
\end{minipage}
&
\begin{minipage}[t]{0.48\columnwidth}
\centering
\includegraphics[width=0.9\linewidth]{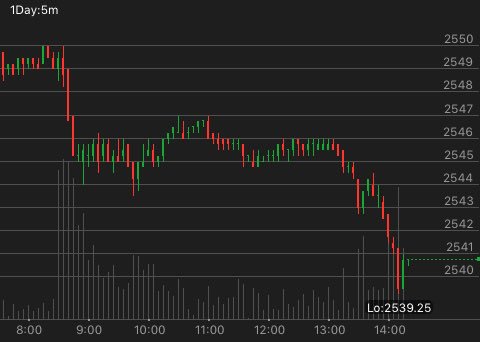} \\
\scriptsize \textbf{(d) Student-only} \\
\scriptsize \emph{Caption:} ``major pullback!! 5pts'' \\
\scriptsize \textbf{GT/T/S:} 1 / 1 / 0 \\
\scriptsize \textbf{Reason:} Sarcasm via exaggeration; student misses numeric reasoning.
\end{minipage}
\end{tabular}
\end{adjustbox}
\caption{\textbf{Qualitative Error Analysis:} Failure cases with GT (Ground Truth), T (Teacher), and S (Student) predictions.}
\label{fig:error_analysis}
\end{figure}

\end{document}